\documentclass[runningheads]{llncs}
\usepackage{graphicx}
\usepackage{tikz}
\usepackage{comment}
\usepackage{amsmath,amssymb} % define this before the line numbering.
\usepackage{color, colortbl}
\definecolor{Gray}{gray}{0.9}
\newcommand{\setParDis}{\setlength {\parskip} {0.2cm} }
\newcommand{\setParDef}{\setlength {\parskip} {0pt} }
\usepackage{epsfig}
\usepackage{amssymb}
\usepackage[utf8]{inputenc}
\usepackage[T1]{fontenc}
\usepackage{url}
\usepackage[ruled,vlined]{algorithm2e}
\usepackage{booktabs}
\usepackage{amsfonts}
\usepackage{nicefrac}
\usepackage{microtype}
\usepackage{wrapfig}
\usepackage{subfigure}
\usepackage{booktabs}
\usepackage{verbatim}
\usepackage{makecell}
\usepackage{cite}
\usepackage{multirow}
\usepackage{bm}
\usepackage{sidecap}
\usepackage{bbding}
\usepackage{pifont}
\newcommand{\bftab}{\fontseries{b}\selectfont}
\usepackage[pagebackref=true,breaklinks=true,colorlinks,citecolor=blue, bookmarks=false]{hyperref}

\begin{document}
\pagestyle{headings}
\mainmatter

\title{Rethinking Confidence Calibration for \\ Failure Prediction}

%\begin{comment}
% \titlerunning{ECCV-22 submission ID \ECCVSubNumber} 
% \authorrunning{ECCV-22 submission ID \ECCVSubNumber} 
% \author{Anonymous ECCV submission}
\institute{Paper ID \ECCVSubNumber}
%\end{comment}
%******************

% \begin{comment}
\titlerunning{Rethinking Confidence Calibration for Failure Prediction}
\author{Fei Zhu\inst{1,2} \and
Zhen Cheng\inst{1,2} \and
Xu-Yao Zhang\inst{1,2}\thanks{Corresponding author.} \and
Cheng-Lin Liu\inst{1,2}}
\authorrunning{F. Zhu et al.}
\institute{NLPR, Institute of Automation, Chinese Academy of Sciences, Beijing 100190, China \and
University of Chinese Academy of Sciences, Beijing, 100049, China\\
\email{\{zhufei2018, chengzhen2019\}@ia.ac.cn, \{xyz, liucl\}@nlpr.ia.ac.cn}}
% \end{comment}
%******************
\maketitle
 
\begin{abstract}
	Reliable confidence estimation for the predictions is important in many safety-critical applications. However, modern deep neural networks are often overconfident for their incorrect predictions. Recently, many calibration methods have been proposed to alleviate the overconfidence problem. With calibrated confidence, a primary and practical purpose is to detect misclassification errors by filtering out low-confidence predictions (known as failure prediction). In this paper, we find a general, widely-existed but actually-neglected phenomenon that most confidence calibration methods are useless or harmful for failure prediction. We investigate this problem and reveal that popular confidence calibration methods often lead to worse confidence separation between correct and incorrect samples, making it more difficult to decide whether to trust a prediction or not. Finally, inspired by the natural connection between flat minima and confidence separation, we propose a simple hypothesis: flat minima is beneficial for failure prediction. We verify this hypothesis via extensive experiments and further boost the performance by combining two different flat minima techniques. Our code is available at \url{https://github.com/Impression2805/FMFP}.

\end{abstract}

\section{Introduction}
Deep neural networks (DNNs), especially vision models, has been widely deployed in risk-sensitive applications such as computer-aided medical diagnosis \cite{miotto2016deep, esteva2017dermatologist}, autonomous driving \cite{janai2017computer, bojarski2016end}, and robotics \cite{leidner2015classifying}. For such applications, besides the prediction accuracy, another crucial requirement is to provide \emph{reliable confidence} for users to make safe decisions. For example, an autonomous driving car should rely more on other sensors or trigger an alarm when the detection network is unable to confidently predict obstructions \cite{janai2017computer}. Another example is the control should be handed over to human doctors when the confidence of a disease diagnosis network is low \cite{miotto2016deep}. Unfortunately, modern DNNs are generally \emph{overconfident} for their predictions, and can easily assign high confidence for misclassified samples \cite{guo2017calibration, hendrycks2017baseline, havasi2020training, corbiere2019addressing}. The overconfident issue makes DNNs models untrustworthy, and therefore brings great concerns when DNNs are deployed in practical applications. 

\begin{figure}[t]
	\begin{center}
		\centerline{\includegraphics[width=0.85\textwidth]{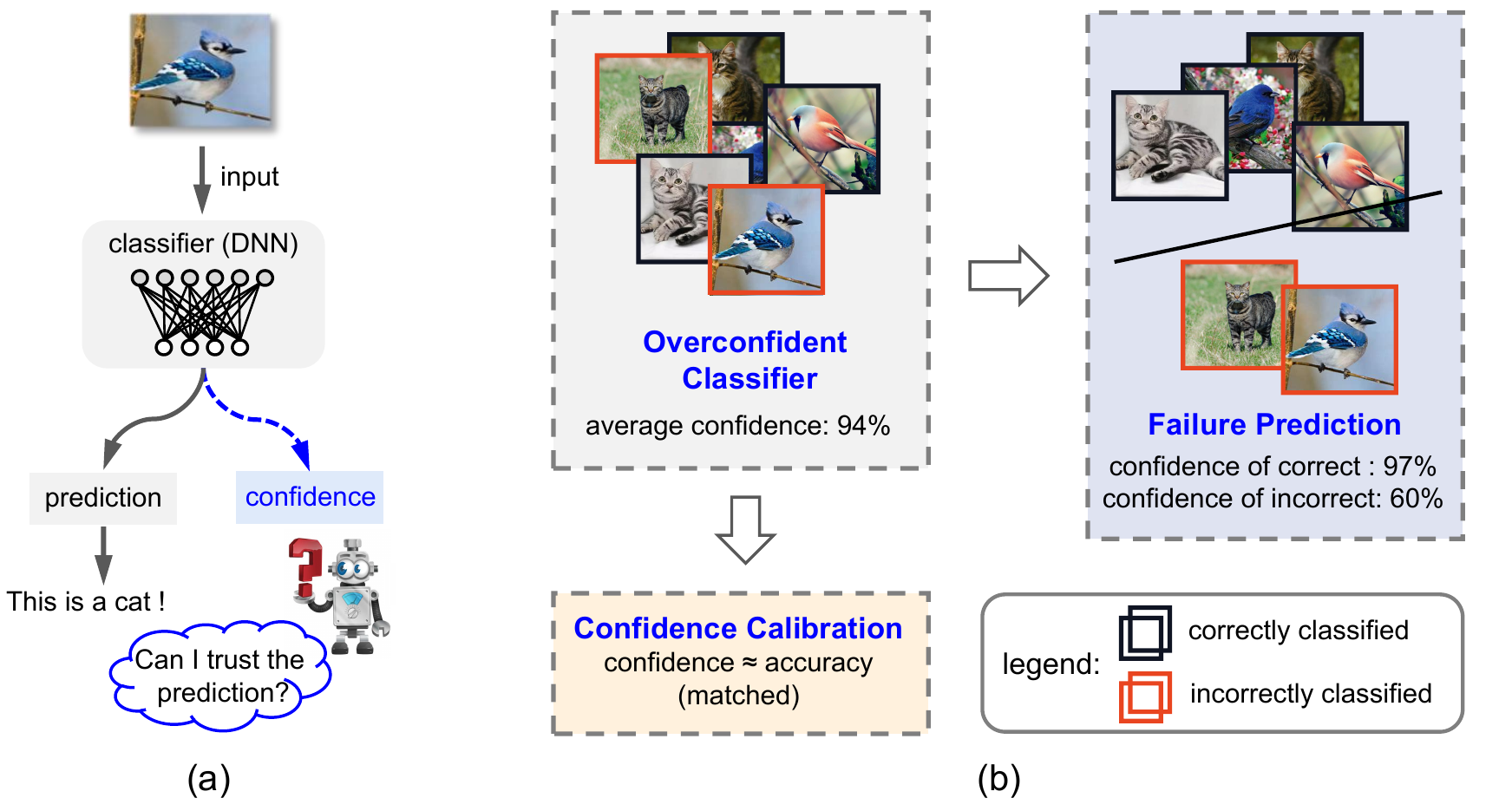}}
		\vskip -0.13 in
		\caption{Confidence calibration aims to reduce the mismatch between a model's confidence and accuracy from the perspective of global statistics, while failure prediction is to distinguish correct from incorrect predictions from the perspective of confidence separability. They both focus on in-distribution data and share the same motivation to provide reliable confidence for trustworthy AI. Therefore, we explore a natural but ignored question: is calibrated confidence useful for failure prediction?}
		\label{figure-1}
	\end{center}
	\vskip -0.35 in
\end{figure}

Recently, many approaches have been developed to alleviate the overconfidence problem by calibrating the confidence, \emph{i.e.}, matching the accuracy and confidence scores to reflect the predictive uncertainty \cite{minderer2021revisiting}. 
Specifically, one category of approaches \cite{pereyra2017regularizing, muller2019does, thulasidasan2019mixup, YunPLS20, XingAZP20, MukhotiKSGTD20, wen2020combining, zhong2021improving, Hebbalaguppe2022CVPR, Liu2022CVPR} aim to learn well-calibrated models during training. For instance, mixup \cite{thulasidasan2019mixup}, label smoothing \cite{muller2019does} and focal loss \cite{MukhotiKSGTD20} have been demonstrated to be effective for confidence calibration.
Another class of approaches \cite{guo2017calibration, RahimiSC0B20, KullFF17, ShehzadBFARKK20, gupta2020calibration, patel2020multi} use post-processing techniques to calibrate DNNs. The most famous post-processing calibration method is temperature scaling \cite{guo2017calibration} which learns a single scalar parameter to calibrate the probabilities.

In this paper, we study a natural but ignored question: \emph{can we use calibrated confidence to detect misclassified samples by filtering out low-confidence predictions}? This, perhaps, is the most direct and practical way to evaluate the quality of the uncertainty. Actually, this problem is studied in the literature as \emph{failure prediction} (also known as misclassification detection or selective classification) \cite{hendrycks2017baseline, corbiere2019addressing, GeifmanE17}, whose purpose is to determine whether the prediction yielded by a classifier is correct or incorrect. Note that failure prediction aims to detect the erroneously classified natural example from seen class (\emph{e.g.}, misclassified samples in test dataset), which is different from the widely studied out-of-distribution (OOD) detection \cite{hendrycks2019deep, LiangLS18, lee2018simple} that focuses on judging whether an input sample is from unseen classes.
Compared with confidence calibration and OOD detection, failure prediction is far less explored in the literature. 

\begin{figure*}[t]
	\begin{center}
		\vskip -0.1 in
		\centerline{\includegraphics[width=\textwidth]{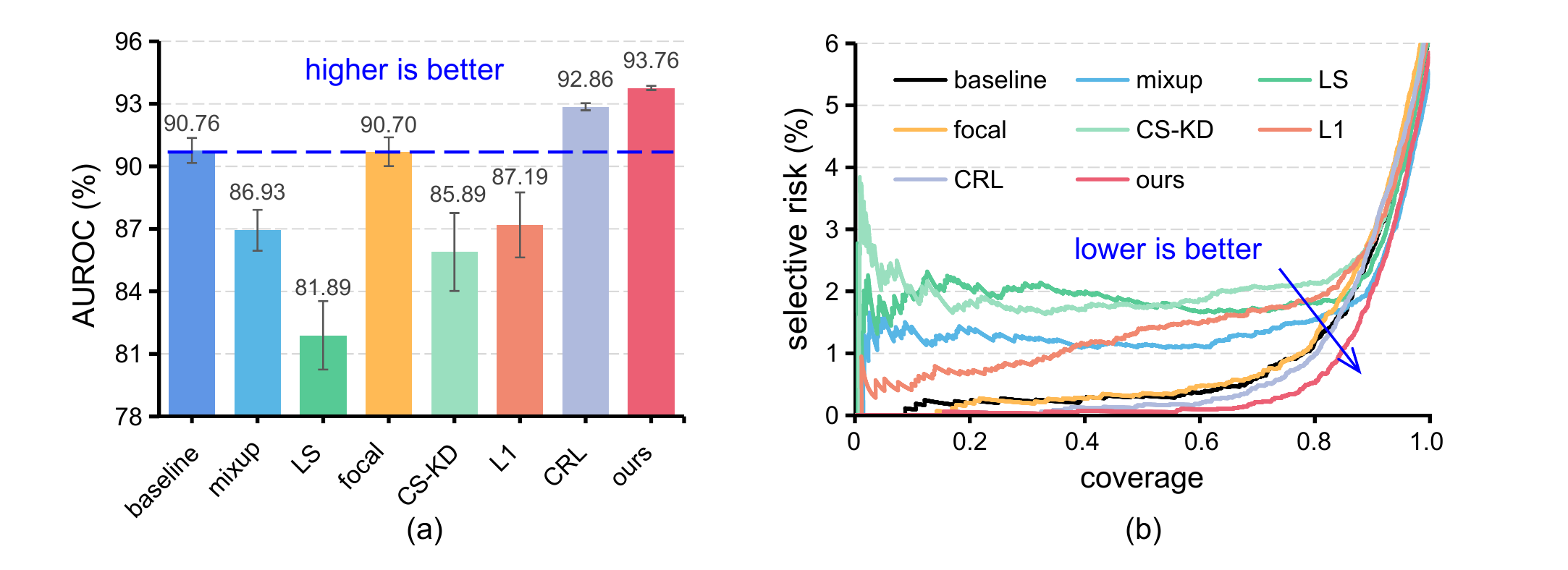}}
		\vskip -0.17 in
		\caption{A comparison of (a) AUROC and (b) risk-coverage curves. We observed that many popular confidence calibration methods are useless or harmful for failure prediction. We propose a simple flat minima based method that can outperform the state-of-the-art failure prediction method CRL \cite{MoonKSH20}. ResNet110 \cite{HeZRS16} on CIFAR-10 \cite{krizhevsky2009learning}.}
		\label{figure-2}
	\end{center}
	\vskip -0.4 in
\end{figure*}
As shown in Fig.~\ref{figure-1}, confidence calibration and failure prediction both focus on the confidence of in-distribution data and share the same motivation that enables the model to provide reliable confidence to make safe decisions. Therefore, common wisdom in the community suggests that calibrated confidence could be useful for failure prediction. However, we find a surprising pathology: many popular confidence calibration methods (including both training-time \cite{pereyra2017regularizing, muller2019does, thulasidasan2019mixup, YunPLS20, HendrycksMCZGL20, XingAZP20} and post-processing \cite{guo2017calibration} calibration methods) are more of a hindrance than a help for failure prediction, as illustrated in Fig.~\ref{figure-2}. Empirical study shows that those methods often reduce overconfidence by simply aligning the accuracy and average confidence. Such calibration could lead to worse separability between correct and misclassified samples, which is harmful for failure prediction. Consequently, one can not effectively detect misclassified samples by filtering out low-confidence predictions based on the calibrated confidence.

Finally, how can we improve the failure prediction performance of DNNs? Intuitively, failure prediction requires better discrimination between the confidence of correct and incorrect samples, which would increase the difficulty of changing the correct samples to be incorrect due to the larger confidence margins. Interestingly, this is closely related to the notion of ``\emph{flatness}'' in DNNs, which reflects how sensitive the correct samples become misclassified when perturbing the model parameters \cite{huang2020understanding, izmailov2018averaging, foret2020sharpness}. Inspired by the natural connection between flat minima and confidence separation, we propose a simple hypothesis: \emph{flat minima is beneficial for failure prediction}. We verify this hypothesis by extensive experiments and propose a simple and effective technique that combines different kinds of flat methods to achieve state-of-the-art performance on failure prediction. 

\setParDis
\noindent
\textbf{Contributions.} Motivated by the widely confirmed confidence calibration effect of recently proposed techniques, we rethink the confidence reliability by evaluating them on the challenging and practical failure prediction task. Surprisingly, we find that they often have negative effect on failure prediction. 
From a detailed analysis, we identify a compounding less-separability effect of training-time calibration methods \cite{thulasidasan2019mixup, muller2019does, LinGGHD20, YunPLS20, joo2020revisiting}, and further find that failure prediction can not be improved by post-hoc calibration strategies like temperature scaling \cite{guo2017calibration}. 
Finally, inspired by the connection between flat minima \cite{huang2020understanding, izmailov2018averaging, foret2020sharpness} and confidence separation, we propose to find flat minima to significantly reduce the confidence of misclassified samples while maintaining the confidence of correct samples. Extensive experiments show the strong performance of our method on both failure prediction and confidence calibration.
\setParDef

\section{Problem Formulation and Background}
Considering the multi-class classification problem, we assume a sample $(\bm{x}, y)$ is drawn \emph{i.i.d.} from an unknown joint distribution over $\mathcal{X} \times \mathcal{Y}$ where $\mathcal{X} = \mathbb{R}^d$ donates the feature space and $\mathcal{Y} = \{1,2,...,k\}$ is a label space. Utilizing a standard softmax function, a deep neural network classifier $f: \mathcal{X} \rightarrow \mathcal{Y}$ produces a probability distribution over $k$ classes. Specifically, given an input $\bm{x}$, $f$ produces the predicted class probabilities $\hat{\bm{p}} = \hat{P}(y|\bm{x}, \bm{\theta})$, where $\bm{\theta}$ is the parameters of the classification model. With these probabilities, $\hat{y} = \arg\max_{y \in \mathcal{Y}} \hat{P}(y|\bm{x}, \bm{\theta})$ can be returned as the predicted class and the associated probability $\hat{p} = \max_{y \in \mathcal{Y}} \hat{P}(y|\bm{x}, \bm{\theta})$, \emph{i.e.}, the maximum class probability, can be viewed as the predicted confidence.

\subsection{Confidence Calibration}
\textbf{Definitions and notation.} Intuitively, the predictive confidence of a well-calibrated model could be indicative of the actual likelihood of correctness \cite{guo2017calibration}. For example, if a calibrated model predicts a set of inputs $\bm{x}$ to be class $y$ with $40\%$ probability, then we expect $40\%$ of the inputs indeed belong to class $y$. Formally, a model is perfectly calibrated if \cite{guo2017calibration, kumar2019verified}:
\begin{equation}
	\label{eq1}
	\begin{aligned}
	P(\hat{y}=y|\hat{p}=p^{\ast}) = p^{\ast}, \forall p^{\ast} \in [0,1].
\end{aligned}
\end{equation}
The most commonly used calibration metric is the Expected Calibration Error (ECE) \cite{Naeini2015ObtainingWC}, which approximates the miscalibration by binning the confidence in $[0,1]$ under $M$ equally-spaced intervals \emph{i.e.,} $\{B_m\}_{m=1}^M$. Then the miscalibration is estimated by taking the expectation of the mismatch between the accuracy and averaged confidence in each bin: $\text{ECE} = \sum_{m=1}^M \frac{\left|B_{m}\right|}{n} \left|\text{acc}(B_{m})-\text{avgConf}(B_{m})\right|$,
where $n$ is the number of all samples. Alternatives to ECE include the negative log likelihood (NLL) and brier score \cite{brier1950verification}.

\setParDis
\noindent
\textbf{Improving calibration.} Many strategies have been proposed to address the miscalibration of modern DNNs. (1) One category of approaches \cite{pereyra2017regularizing, muller2019does, thulasidasan2019mixup, YunPLS20, XingAZP20, MukhotiKSGTD20, wen2020combining, zhong2021improving} aim to learn well-calibrated models during training. For example, several works \cite{thulasidasan2019mixup, mixCalibration, howmixup} found that the predicted scores of DNNs trained with mixup \cite{zhang2018mixup} are better calibrated. Muller \emph{et al}. \cite{muller2019does} showed the favorable calibration effect of label smoothing. Mukhoti \emph{et~al}. \cite{MukhotiKSGTD20} demonstrated that focal loss \cite{LinGGHD20} can automatically learn well-calibrated models. CS-KD \cite{YunPLS20} calibrates overconfident predictions by penalizing the predictive distribution between the samples within the same class. Recently, Joo \emph{et~al}. \cite{joo2020revisiting} explored the effect of explicit regularization strategies (\emph{e.g.}, $L_p$ norm in the logits space) for calibration.
(2) Another class of approaches \cite{guo2017calibration, RahimiSC0B20, KullFF17, ShehzadBFARKK20, gupta2020calibration, patel2020multi} rescal the predictions in a post-hoc manner. Among them, temperature scaling \cite{guo2017calibration} is a effective and simple technique, which has inspired various post-processing approaches \cite{KullPKFSF19, Mozafari2019UnsupervisedTS, ShehzadBFARKK20}. 

\noindent
\textbf{Empirical studies of calibration.}
In addition to the calibration strategies, there have been some empirical studies on calibration. Guo \emph{et al.} \cite{guo2017calibration} observed that larger networks tend to be less calibrated, even as classification accuracy is improved.
Ovadia \emph{et al}. \cite{ovadia2019can} studied the calibration under distribution shift and empirically found the generally existing performance drop of different calibration methods under distribution shift. More recently, Wang \emph{et al}. \cite{RethinkingCalibration} found that it is harder to further calibrate the model with temperature scaling if it has been trained with regularization methods. Minderer \emph{et al}. \cite{minderer2021revisiting} found that the most recent non-convolutional models \cite{dosovitskiy2020image, tolstikhin2021mlp} are well-calibrated, suggesting that architecture is a major factor of calibration performance. Differently from those works, we rethink the confidence calibration for failure prediction.
\setParDef

\subsection{Failure Prediction}
\textbf{Definitions and notation.} Failure prediction, also known as misclassification detection \cite{hendrycks2017baseline} or ordinal ranking \cite{MoonKSH20}, focus on distinguishing incorrect from correct predictions based on their confidence ranking. Intuitively, if the associated confidence of each misclassified sample is lower than that of any correctly classified samples, we can successfully predict each error made by the classification model at inference time. Formally, an optimal ordinal ranking model should reflect the following relationship for every two samples $(\bm{x}_i, y_i)$ and $(\bm{x}_j, y_j)$:
\begin{equation}
\label{eq2}
\begin{aligned}
	\kappa(\hat{\bm{p}}_i|\bm{x_i},\bm{\theta}) \ge \kappa(\hat{\bm{p}}_j|\bm{x_j},\bm{\theta}) \Longleftrightarrow P(\hat{y}_i=y_i|\bm{x_i}) \ge P(\hat{y}_j=y_j|\bm{x_j}),
\end{aligned}
\end{equation}
where $\kappa$ denotes a confidence-rate function (\emph{e.g.}, the maximum class probability) that assess the degree of confidence of the predictions. Then, with a predefined threshold $\delta \in \mathbb{R}^{+}$, the users can reject the erroneously classification results based on the following decision function $g$:
\begin{equation}
	\label{eq3}
		g(\bm{x}) = \left\{ 
			\begin{aligned}
			\text{accept}, &~\text{if} ~~\kappa(\bm{x}) \ge \delta, \\
		    \text{reject}, &~\text{otherwise}.
		\end{aligned}
	\right.
\end{equation}
Common metrics for failure prediction are the risk-coverage curve (AURC), the normalized AURC (E-AURC) \cite{GeifmanE17, MoonKSH20}, the false positive rate at $95\%$ true positive rate (FPR-95\%TPR), and the area under the receiver operating characteristic curve (AUROC). In addition, there are some other metrics \cite{hendrycks2017baseline} to imply how the correct and incorrect predictions are separated.

\setParDis
\noindent
\textbf{Improving failure prediction.} For DNNs, Hendrycks \emph{et al}. \cite{hendrycks2017baseline} firstly established a \textbf{baseline} by using maximum softmax probability. Trust-Score \cite{Jiang2018ToTO} adopts the similarity between the classifier and a nearest-neighbor classifier as a confidence measure. However, it lacks of practicality and scalability in high-dimensional spaces. 
Some works \cite{corbiere2019addressing, luo2021learning} formulate the failure prediction as a supervised binary classification problem. For example, ConfidNet \cite{corbiere2021confidence, corbiere2019addressing} train auxiliary models to predict confidence by learning the misclassified samples in training set. However, they may fail when the model has a high training accuracy, in which few or even no misclassified examples will exist in the training set. CRL \cite{MoonKSH20} improves failure prediction by regularizing the model to learn an ordinal ranking relationship based on the historical correct rate during training. More recently, Zhu \emph{et al}. established a unified confidence estimation method named classAug \cite{zhu2022learning} that can improve calibration, failure prediction and OOD detection. To the best of our knowledge, there are only a few works for failure prediction and only CRL \cite{MoonKSH20} and classAug \cite{zhu2022learning} can outperform the baseline \cite{hendrycks2017baseline}. 

\setParDis
\noindent
\textbf{Why revisit confidence calibration for failure prediction?}
(1) Confidence calibration and failure prediction both focus on the confidence of in-distribution data and share the same motivation to provide reliable confidence for making safer decisions. From a practical perspective, with a calibrated classifier in hand, one natural way to verify its trustworthiness is to filter out predictions with low confidence. 
(2) Confidence calibration has drawn significant attention from the machine learning community, including improving calibration \cite{muller2019does, thulasidasan2019mixup, YunPLS20, guo2017calibration}, empirical studies of calibration \cite{guo2017calibration, ovadia2019can, wen2020combining, minderer2021revisiting} and measures of calibration \cite{nixon2019measuring, vaicenavicius2019evaluating, gupta2020calibration, brier1950verification}. However, there are few works for failure prediction, which is a practical, important, yet somewhat under-appreciated area of research. Revisiting calibration from the perspective of failure prediction not only helps understand calibration but also benefits the investigation of failure prediction.
\setParDef

\begin{table}[!t]
	\caption{Failure prediction performance on CIFAR-10 and CIFAR-100 datasets. AURC and E-AURC values are multiplied by $10^3$, and all remaining values are percentage.}
	\vskip -0.22in
	\label{table-1}
	\begin{center}
	\renewcommand\tabcolsep{8pt}
		\begin{small}
			\newcommand{\tabincell}[2]{\begin{tabular}{@{}#1@{}}#2\end{tabular}}
			\scalebox{0.57}{
				\renewcommand{\arraystretch}{1}
				\begin{tabular}{llcccccc}
					\toprule
					\multicolumn{8}{c}{\textbf{CIFAR-10}}  \\ \midrule
					\textbf{Network} 
					& \textbf{Method}&
					\tabincell{c}{\textbf{AURC} \\ \textbf{($\downarrow$)}} & \tabincell{c}{\textbf{E-AURC} \\ \textbf{($\downarrow$)}}&
					\tabincell{c}{\textbf{FPR-95\%} \\ \textbf{TPR($\downarrow$)}} &
					\tabincell{c}{\textbf{AUROC} \\ \textbf{($\uparrow$)}} & \tabincell{c}{\textbf{AUPR-} \\ \textbf{Success($\uparrow$)}} & \tabincell{c}{\textbf{AUPR-} \\ \textbf{Error($\uparrow$)}} \\
					\midrule
					\multirow{6}{*}{ResNet110}
					& baseline \cite{hendrycks2017baseline} & \bftab{9.94$\pm$1.29} & \bftab{7.94$\pm$1.23} & 45.01$\pm$2.55 & \bftab{90.76$\pm$0.60} & \bftab{99.16$\pm$0.13} & 43.66$\pm$1.86 \\
					& mixup \cite{thulasidasan2019mixup} & 15.05$\pm$0.09 & 13.31$\pm$0.29 & \bftab{41.18$\pm$2.54} & 86.93$\pm$0.98 & 98.60$\pm$0.03 & 43.19$\pm$3.52 \\
					& LS \cite{muller2019does} & 24.50$\pm$2.73 & 22.57$\pm$2.77 & 44.76$\pm$3.76 & 81.89$\pm$1.64 & 97.63$\pm$0.29 & 39.57$\pm$1.00 \\
					& Focal \cite{MukhotiKSGTD20} & 10.70$\pm$1.32 & 8.30$\pm$1.25 & 45.85$\pm$0.75 & 90.70$\pm$0.69 & 99.12$\pm$0.13 & \bftab{44.32$\pm$1.40} \\
					& CS-KD \cite{YunPLS20} & 18.02$\pm$3.47 & 15.81$\pm$3.43 & 43.98$\pm$3.86 & 85.89$\pm$1.87 & 98.33$\pm$0.36 & 41.17$\pm$4.09 \\
					& L1 \cite{YunPLS20} & 14.21$\pm$1.93 & 12.11$\pm$1.92 & 45.71$\pm$0.60 & 87.19$\pm$1.56 & 98.72$\pm$0.20 & 41.93$\pm$1.51 \\

					\midrule
					\multirow{6}{*}{WRNet}
					& baseline \cite{hendrycks2017baseline}  & \bftab{4.89$\pm$0.25} & \bftab{3.93$\pm$0.22} & 32.85$\pm$0.36 & \bftab{93.24$\pm$0.15} & \bftab{99.59$\pm$0.02} & \bftab{43.38$\pm$1.12} \\
					& mixup \cite{thulasidasan2019mixup} & 6.24$\pm$0.79 & 5.52$\pm$0.80 & \bftab{31.27$\pm$0.63} & 91.18$\pm$0.66 & 99.43$\pm$0.08 & 41.86$\pm$0.79 \\
					& LS \cite{muller2019does} & 16.69$\pm$2.86 & 15.69$\pm$2.86 & 34.24$\pm$2.84 & 85.30$\pm$1.79 & 98.38$\pm$0.29 & 43.19$\pm$1.27 \\
					& Focal \cite{MukhotiKSGTD20} & 6.91$\pm$0.56 & 5.83$\pm$0.55 & 35.43$\pm$2.52 & 91.86$\pm$0.66 & 99.39$\pm$0.06 & 43.11$\pm$0.37 \\
					& CS-KD \cite{YunPLS20} & 10.29$\pm$0.35 & 9.22$\pm$0.38 & 38.65$\pm$2.67 & 88.13$\pm$0.74 & 99.04$\pm$0.04 & 38.88$\pm$0.66 \\
					& L1 \cite{YunPLS20} & 7.01$\pm$1.63 & 5.99$\pm$1.66 & 34.19$\pm$0.78 & 91.08$\pm$1.91 & 99.38$\pm$0.17 & 42.25$\pm$1.51 \\

					\midrule

					\multirow{6}{*}{DenseNet}
					& baseline \cite{hendrycks2017baseline} & \bftab{6.2$\pm$0.29} & \bftab{4.66$\pm$0.27} & 38.20$\pm$2.48  & \bftab{92.87$\pm$0.44} & \bftab{99.51$\pm$0.03} & 43.74$\pm$2.44 \\
					& mixup \cite{thulasidasan2019mixup} & 8.57$\pm$0.54 & 7.08$\pm$0.51 & \bftab{37.88$\pm$3.73} & 91.17$\pm$0.84 & 99.26$\pm$0.05 & \bftab{44.35$\pm$}1.16 \\
					& LS \cite{muller2019does} & 19.35$\pm$2.29 & 17.67$\pm$2.19 & 40.20$\pm$2.10 & 84.45$\pm$1.45 & 98.15$\pm$0.23 & 41.35$\pm$1.18\\
					& Focal \cite{MukhotiKSGTD20} & 7.17$\pm$0.28 & 5.35$\pm$0.19 & 41.75$\pm$0.88 & 92.31$\pm$0.17 & 99.44$\pm$0.02 & 43.27$\pm$2.04  \\
					& CS-KD \cite{YunPLS20} & 13.55$\pm$1.02 & 11.64$\pm$0.92 & 43.11$\pm$3.75 & 88.02$\pm$0.46 & 98.77$\pm$0.10 & 40.08$\pm$4.30\\
					& L1 \cite{YunPLS20} &8.34$\pm$1.24 &6.94$\pm$1.07 &37.02$\pm$0.97 &91.40$\pm$0.63 &99.27$\pm$0.13 & 44.19$\pm$0.55\\

					\midrule
					\multirow{6}{*}{ConvMixer}
					& baseline \cite{hendrycks2017baseline} & \bftab{8.33$\pm$1.44} & \bftab{6.29$\pm$1.30} & 42.32$\pm$3.26 & \bftab{92.02$\pm$0.96} & \bftab{99.34$\pm$0.14} & 43.80$\pm$1.49 \\
					& mixup \cite{thulasidasan2019mixup} & 9.87$\pm$0.14 & 8.40$\pm$0.25 & \bftab{37.57$\pm$2.10} & 90.25$\pm$0.81 & 99.12$\pm$0.02 & \bftab{45.01$\pm$2.40} \\
					& LS \cite{muller2019does} & 18.45$\pm$1.12 & 16.41$\pm$1.08 & 40.99$\pm$1.91 & 86.01$\pm$0.24 & 98.27$\pm$0.11 & 43.32$\pm$0.70 \\
					& Focal \cite{MukhotiKSGTD20} & 9.59$\pm$1.02 & 7.17$\pm$1.03 & 46.18$\pm$1.66 & 91.32$\pm$0.87 & 99.24$\pm$0.11 & 44.03$\pm$0.57 \\
					& CS-KD \cite{YunPLS20} & 13.62$\pm$0.86 & 11.69$\pm$0.93 & 43.06$\pm$0.35 & 88.02$\pm$0.53 & 98.77$\pm$0.10 & 41.35$\pm$0.78 \\
					& L1 \cite{YunPLS20} &12.92$\pm$2.41 & 11.07$\pm$2.36 & 41.39$\pm$0.91 & 88.35$\pm$1.57 & 98.83$\pm$0.25 & 42.91$\pm$1.60 \\
					
					\bottomrule
					\toprule
					\multicolumn{8}{c}{\textbf{CIFAR-100}}  \\ \midrule
					\textbf{Network} 
					& \textbf{Method}&
					\tabincell{c}{\textbf{AURC} \\ \textbf{($\downarrow$)}} & \tabincell{c}{\textbf{E-AURC} \\ \textbf{($\downarrow$)}}&
					\tabincell{c}{\textbf{FPR-95\%} \\ \textbf{TPR($\downarrow$)}} &
					\tabincell{c}{\textbf{AUROC} \\ \textbf{($\uparrow$)}} & \tabincell{c}{\textbf{AUPR-} \\ \textbf{Success($\uparrow$)}} & \tabincell{c}{\textbf{AUPR-} \\ \textbf{Error($\uparrow$)}}\\
					\midrule
					\multirow{6}{*}{ResNet110}
					& baseline \cite{hendrycks2017baseline} & \bftab{93.90$\pm$2.37} & \bftab{50.88$\pm$2.03} & 66.02$\pm$1.53 & \bftab{85.00$\pm$0.35} & \bftab{93.42$\pm$0.28} & \bftab{66.54$\pm$0.29} \\
					& mixup \cite{thulasidasan2019mixup} & 95.03$\pm$2.77 & 57.57$\pm$3.33 & \bftab{63.68$\pm$0.36} & 84.03$\pm$0.80 & 92.65$\pm$0.40 & 64.55$\pm$1.46 \\
					& LS \cite{muller2019does} & 111.18$\pm$0.36 & 69.46$\pm$1.08 & 63.93$\pm$0.65 & 82.85$\pm$0.34 & 91.00$\pm$0.11 & 65.19$\pm$0.32 \\
					& Focal \cite{MukhotiKSGTD20} & 96.60$\pm$2.81 & 52.97$\pm$1.96 & 66.60$\pm$0.64 & 84.22$\pm$0.17 & 93.16$\pm$0.28 & 65.23$\pm$0.53 \\
					& CS-KD \cite{YunPLS20} & 100.68$\pm$2.83 & 58.89$\pm$1.94 & 66.15$\pm$1.69 & 83.98$\pm$0.21 & 92.39$\pm$0.27 & 65.09$\pm$0.55 \\
					& L1 \cite{YunPLS20} & 119.49$\pm$4.04 & 71.47$\pm$3.31 & 65.78$\pm$0.96 & 82.73$\pm$0.51 & 90.55$\pm$0.44 & 66.44$\pm$0.59 \\
	
					\midrule
					\multirow{6}{*}{WRNet}
					& baseline \cite{hendrycks2017baseline} & 51.97$\pm$1.74 & \bftab{30.18$\pm$1.09} & 59.19$\pm$0.20 & \bftab{87.75$\pm$0.30} & \bftab{96.38$\pm$0.14} & \bftab{62.99$\pm$0.98} \\
					& mixup \cite{thulasidasan2019mixup} & \bftab{50.54$\pm$0.98} & 32.00$\pm$0.92 & \bftab{57.29$\pm$1.57} & 87.47$\pm$0.43 & 96.21$\pm$0.10 & 61.72$\pm$1.25 \\
					& LS \cite{muller2019does} & 58.29$\pm$3.46 & 36.69$\pm$2.66 & 58.47$\pm$0.98 & 86.76$\pm$0.19 & 95.59$\pm$0.34 & 61.95$\pm$0.72 \\
					& Focal \cite{MukhotiKSGTD20} & 54.54$\pm$1.44 & 32.40$\pm$1.28 & 61.87$\pm$1.06 & 86.89$\pm$0.22 & 96.12$\pm$0.16 & 60.56$\pm$0.65 \\
					& CS-KD \cite{YunPLS20} & 58.30$\pm$1.29 & 35.62$\pm$0.68 & 60.44$\pm$0.95 & 86.98$\pm$0.19 & 95.70$\pm$0.09 & 62.23$\pm$1.17 \\
					& L1 \cite{YunPLS20} & 61.63$\pm$0.46 & 38.78$\pm$0.45 & 59.48$\pm$2.07 & 86.21$\pm$0.28 & 95.31$\pm$0.05 & 62.55$\pm$0.77 \\

					\midrule
					
					\multirow{6}{*}{DenseNet}
					& baseline \cite{hendrycks2017baseline} & 67.41$\pm$0.67 & \bftab{35.82$\pm$0.54} & \bftab{61.55$\pm$2.01} & \bftab{86.46$\pm$0.31} & \bftab{95.55$\pm$0.05} & \bftab{65.60$\pm$1.26} \\
					& mixup \cite{thulasidasan2019mixup} & \bftab{64.84$\pm$4.26} & 37.06$\pm$2.74 & 62.94$\pm$2.55 & 86.26$\pm$0.63 & 95.47$\pm$0.36 & 62.64$\pm$0.81 \\
					& LS \cite{muller2019does} & 76.24$\pm$2.31 & 44.40$\pm$1.01 & 62.41$\pm$0.51 & 85.32$\pm$0.10 & 94.47$\pm$0.15 & 63.59$\pm$0.43 \\
					& Focal \cite{MukhotiKSGTD20} & 73.43$\pm$2.05 & 40.70$\pm$1.40 & 65.67$\pm$1.48 & 85.62$\pm$0.30 & 94.95$\pm$0.19 & 62.76$\pm$0.61 \\
					& CS-KD \cite{YunPLS20} & 75.22$\pm$1.02 & 41.38$\pm$1.28 & 62.75$\pm$ 0.85 & 86.20$\pm$0.40 & 94.82$\pm$0.15 & 64.50$\pm$0.30 \\
					& L1 \cite{YunPLS20} &68.73$\pm$1.10 & 40.47$\pm$1.37 &63.39$\pm$0.92 &85.46$\pm$0.50 &95.06$\pm$0.22 &61.90$\pm$0.44\\

					\midrule
					\multirow{6}{*}{ConvMixer}
					& baseline \cite{hendrycks2017baseline} & 76.96$\pm$1.64 & 41.20$\pm$1.64 & \bftab{63.57$\pm$0.52} & \bftab{86.28$\pm$0.18} & 94.81$\pm$0.06 & \bftab{65.39$\pm$0.83} \\
					& mixup \cite{thulasidasan2019mixup} & \bftab{70.87$\pm$2.25} & \bftab{39.38$\pm$2.65} & 63.80$\pm$1.72 & 86.12$\pm$0.75 & \bftab{95.12$\pm$0.75} & 63.71$\pm$0.75 \\
					& LS \cite{muller2019does} & 83.50$\pm$3.88 & 47.37$\pm$2.45 & 66.49$\pm$1.50 & 84.69$\pm$0.62 & 94.03$\pm$0.33 & 62.87$\pm$1.24 \\
					& Focal \cite{MukhotiKSGTD20} & 83.79$\pm$2.03 & 45.51$\pm$0.88 & 66.26$\pm$1.50 & 85.17$\pm$0.34 & 94.24$\pm$0.12 & 64.36$\pm$1.76 \\
					& CS-KD \cite{YunPLS20} & 74.02$\pm$1.18 & 42.18$\pm$ 0.56 & 64.83$\pm$1.46 & 85.44$\pm$0.14 & 94.76$\pm$0.09 & 62.64$\pm$0.69 \\
					& L1 \cite{YunPLS20} &82.68$\pm$0.35 & 46.62$\pm$0.50 & 65.03$\pm$0.44 & 85.02$\pm$0.14 & 94.12$\pm$0.05 & 64.01$\pm$0.42 \\
					
					\bottomrule
			\end{tabular}}
		\end{small}
	\end{center}
	\vskip -0.3in
\end{table}

\section{Does Calibration Help Failure Prediction?}
In recent years, there is a surge of research focused on alleviating the overconfidence problem of modern DNNs. As shown by many empirical results, existing methods do help the calibration of DNNs. In this section, we empirically investigate the reliability of the calibrated confidence for failure prediction.

\subsection{Experimental Setup}
\noindent
\textbf{Datasets and network architectures.} 
We thoroughly conduct experiments on benchmark datasets CIFAR-10 and CIFAR-100 \cite{krizhevsky2009learning}, and large-scale ImageNet \cite{deng2009imagenet} dataset. In terms of network architectures, we consider a range of models: PreAct-ResNet110 \cite{HeZRS16}, WideResNet \cite{zagoruyko2016wide}, DenseNet \cite{HuangLMW17} and more recent architecture ConvMixer \cite{trockman2022patches} for experiments on CIFAR-10 and CIFAR-100. For ImageNet, we used a ResNet-18 \cite{he2016deep} model. Due to space limitation, we provide the results of more networks like MobileNet \cite{howard2017mobilenets}, EfficientNet \cite{tan2019efficientnet} and dataset like Tiny-ImageNet \cite{yao2015tiny} in the the supplementary material.

\setParDis
\noindent
\textbf{Evaluation metrics.} We adopt the standard metrics in \cite{hendrycks2017baseline} and \cite{GeifmanE17, MoonKSH20} to measure failure prediction: AURC, E-AURC, AUROC, FPR-95\%TPR, AUPR-Success and AUPR-Error. Lower values of AURC, E-AURC, FPR-95\%TPR and higher values of AUROC, AUPR-Success, AUPR-Error
indicate better failure prediction ability. Supplementary material provides definitions of these metrics. 

\noindent
\textbf{Implementation details.} All models are trained using SGD with a momentum of 0.9, an initial learning rate of 0.1, and a weight decay of 5e-4 for 200 epochs with the mini-batch size of 128 on CIFAR-10 and CIFAR-100. The learning rate is reduced by a factor of 10 at 80, 130, and 170 epochs. We randomly sample $10\%$ of training samples as a validation dataset for each task because it is a requirement for post-calibration methods like temperature scaling \cite{guo2017calibration}. For each experiment, the mean and standard deviation over three random runs are reported. 

\begin{figure*}[t]
	\begin{center}
		\centerline{\includegraphics[width=\textwidth]{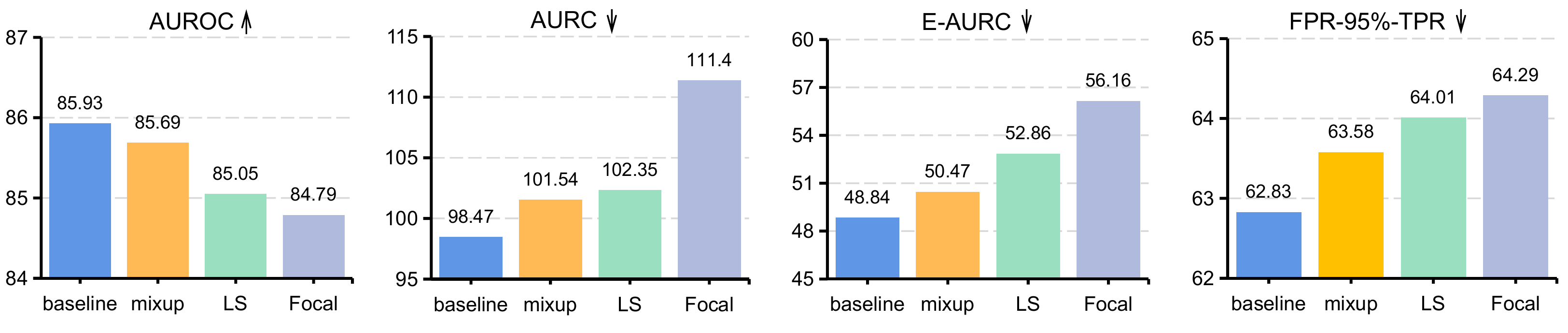}}
		\vskip -0.1 in
		\caption{Large-scale experiments on ImageNet. AURC and E-AURC values are multiplied by $10^3$ for clarity, and all remaining values are percentage.}
		\label{figure-3}
	\end{center}
	\vskip -0.3 in
\end{figure*}
\noindent
\textbf{Evaluated calibration methods.} We evaluate various calibration methods including training-time regularization like mixup \cite{thulasidasan2019mixup}, label-smoothing (LS) \cite{muller2019does}, focal loss \cite{LinGGHD20}, CS-KD \cite{YunPLS20}, $L_1$ norm \cite{joo2020revisiting} and post-hoc method like temperature scaling (TS) \cite{guo2017calibration}. Those methods have been verified to be effective to address the miscalibration problem of DNNs. Particularly, compared with post-hoc methods, the effect of training-time regularization is irreversible. Therefore, we mainly focus on their performance on failure prediction. Supplementary material provides detail introduction and hyperparameter setting of each method.
\setParDef

\subsection{Results and Analysis} 
In our experiments, we confirmed the positive calibration effects of the evaluated methods. For example, on CIFAR-10, with mixup, the ECE (\%) can be reduced from 4.14 to 2.97 for ResNet110 and from 2.96 to 1.39 for DenseNet; with focal loss, the ECE (\%) can be reduced from 4.14 to 1.60 for ResNet110 and from 2.96 to 1.36 for DenseNet. These observations are consistent with that in \cite{thulasidasan2019mixup, LinGGHD20}.

\setParDis
\noindent
\textbf{Popular calibration methods can harm failure prediction.}
In practice, users would naturally expect that the calibrated confidence can be used to filter out low-confidence predictions in risk-sensitive applications.
However, if we shift focus to Table~\ref{table-1}, it is evident that those methods generally lead to \textbf{\emph{worse}} failure prediction performance under various metrics. For example, when training with mixup and LS on CIFAR-10/ResNet110, the AUROC ($\uparrow$) drops $3.83$ and $9.07$ percentages,  respectively. And the AURC ($\downarrow$) increases $5.51$ and $14.56$ percentages, respectively. This is counter-intuitive as we expect those methods, which successfully calibrate the confidence, could be useful for failure prediction.

\setParDis
\noindent
\textbf{The same observations generalize to large-scale dataset.} Here we verify our observation that calibration methods often harm failure prediction on ImageNet \cite{deng2009imagenet} dataset, which comprises 1000 classes and over 1.2 million images. We train a ResNet-18 \cite{he2016deep} that achieve a $70.20\%$ top-1 classification accuracy. The results are shown in Fig.~\ref{figure-3}, from which we can observe similar negative effect of calibration methods on failure prediction. More results on other networks, which exhibit similar pattern, can be found in supplementary material.

\noindent
\textbf{Selective risk analysis.} To make intuitive sense of the effect of those calibration methods on failure prediction, Fig.~\ref{figure-2}(b) plots the risk-coverage curve. Specifically, selective risk is the empirical loss or error rate that trust the prediction, while coverage is the probability mass of non-rejected predictions \cite{GeifmanE17, MoonKSH20}. Intuitively, a better failure predictor should have low risk at a given coverage. As can be seen from Fig.~\ref{figure-2}(b), the baseline has the lowest risks compared to other calibration methods, which indicates that using the confidence calibrated by those methods would unfortunately increase the risk when making decisions. 

\begin{figure*}[t]
	\begin{center}
		\centerline{\includegraphics[width=0.95\textwidth]{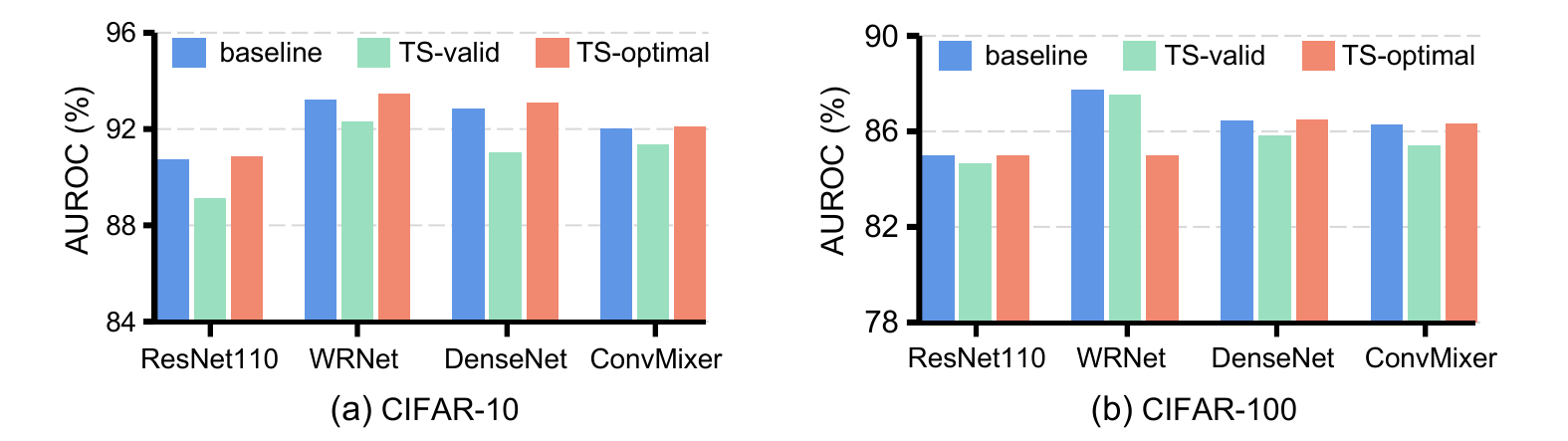}}
		\vskip -0.15 in
		\caption{Temperature scaling can hardly improve failure prediction.}
		\label{figure-5}
	\end{center}
	\vskip -0.35 in
\end{figure*}

\noindent \textbf{Does temperature scaling improve failure prediction?} As a representative post-hoc calibration technique, TS \cite{guo2017calibration} is simple and effective. Specifically, TS calibrates probabilities by learning a single scalar parameter $T$ for all classes on a hold-out validation set. In Fig.~\ref{figure-5}, we show the failure prediction performance of TS on different networks and datasets. Specifically, using validation set and test set to learn the parameter $T$ are donated as \emph{TS-valid} and \emph{TS-optimal}, respectively. By directly using test set, TS-optimal yields the optimal $T$ for failure prediction. As shown in Fig.~\ref{figure-5}, compared with baseline, TS-valid has negative effectiveness while TS-optimal has negligible improvement.
\setParDef

\subsection{Calibration Harms Failure Prediction: A Closer Look}
\begin{wrapfigure}{r}{5.5cm}
	\centering
	\vskip -0.35 in
	\includegraphics[width=5.5cm]{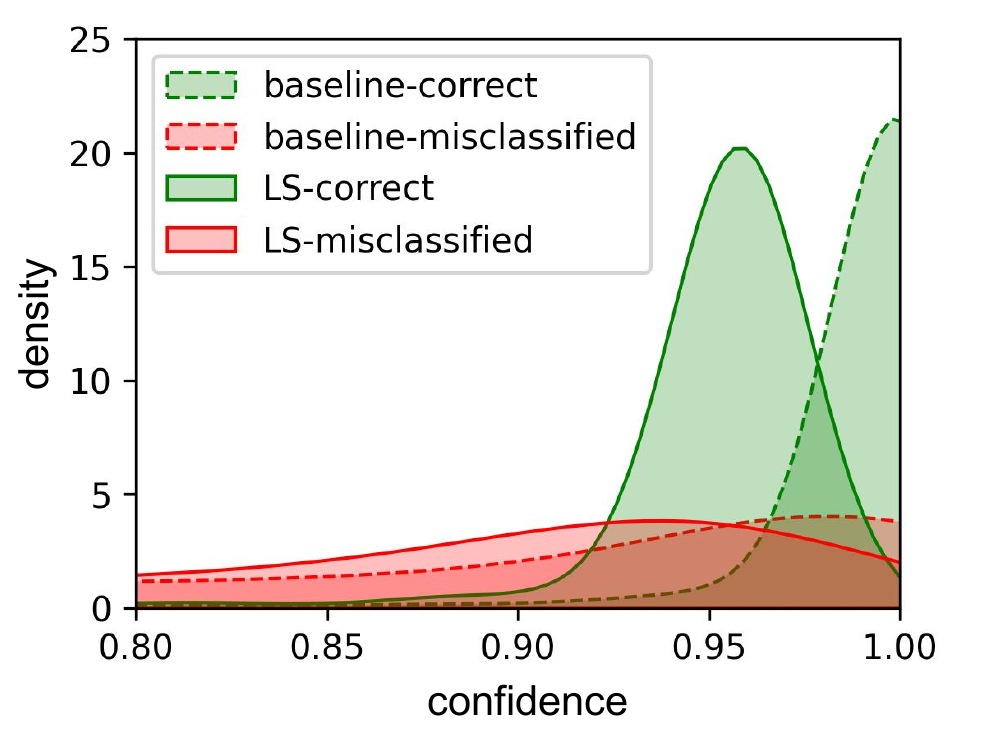}
	\vskip -0.17 in
	\caption{LS results in under-confident correct samples. ResNet110/CIFAR-10.}
	\label{figure-6}
	\vskip -0.25 in
\end{wrapfigure}
\begin{figure*}[t]
	\begin{center}
		\centerline{\includegraphics[width=1.15\textwidth]{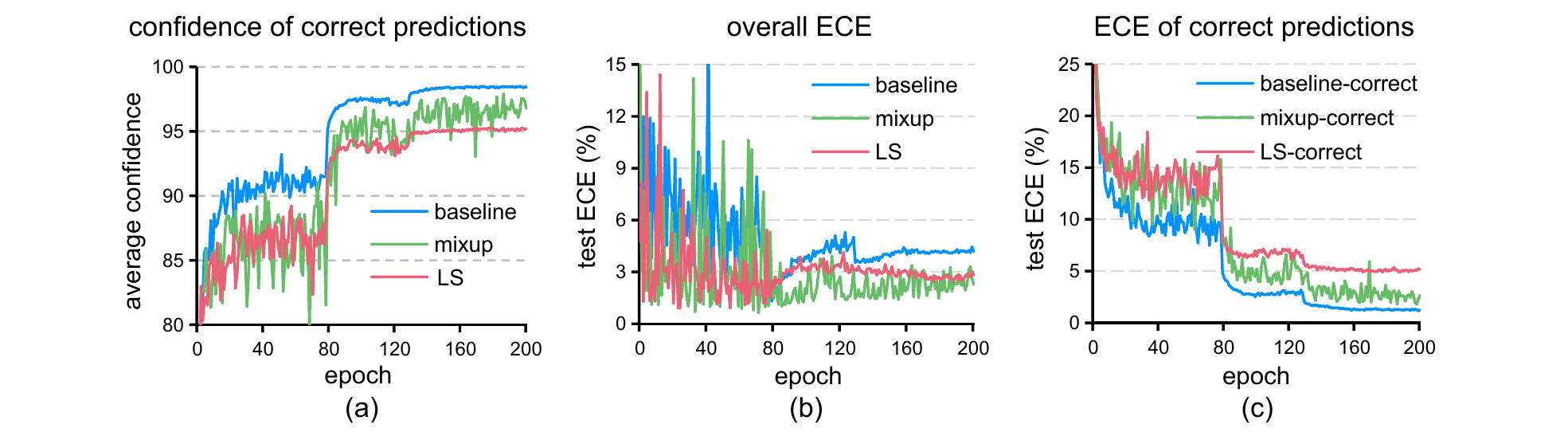}}
		\vskip -0.12 in
		\caption{(a) Comparison of average confidence of correctly classified samples during training. (b) Mixup and LS successfully reduce the overall ECE, (c) but result in worse ECE of correct samples. ResNet110 on CIFAR-10.}
		\label{figure-7}
	\end{center}
	\vskip -0.3 in
\end{figure*}

\textbf{Illustrative experiments.} To empirically understand the negative effect of those calibration methods, Fig.~\ref{figure-6} shows the confidence distribution of test samples. It can be seen that LS leads to a more severe overlap between the confidence of correct and incorrect samples. Fig.~\ref{figure-7}(a) plots the average confidence of correctly classified samples during training, where their confidences are obviously reduced. This can also be seen from Fig.~\ref{figure-7}(b-c), in which mixup and LS lead to better overall ECE but worse ECE of correct samples.
This indicates that those calibration methods yield under-confident correct prediction.
In conclusion, those calibration methods reduce the overconfident of DNNs, but lead to worse separability between correct and misclassified samples, making it hard to detect misclassified samples based on the calibrated confidence.

\setParDis
\noindent
\textbf{Discussion on calibration for failure prediction.} 
The best failure prediction is achieved when correct and wrong predictions are clearly separated according to the confidence level. However, calibration focuses on matching the average accuracy and confidence, and the best ECE score is achieved when correct and wrong predictions are ``mixed'' in the right way such that those with confidence at level $[c,c+w)$ (confidence window in a bin, where $c$ donates the value of confidence) should have a mix of correct and wrong predictions with ratios $c:1-c$. Regularization methods such as mixup \cite{thulasidasan2019mixup}, LS \cite{muller2019does}, focal loss \cite{LinGGHD20}, CS-KD \cite{YunPLS20} and $L_p$ norm \cite{joo2020revisiting}  typically improve calibration by penalizing the confidence of the whole samples to a low level. However, this will lead to undesirable effects: erasing important information about the hardness of samples \cite{shen2020label}, which would introduce an undesired mixing in ranking, and thus result in drops in failure prediction qualities. Nevertheless, this does not mean that a better ECE must lead to worse failure prediction. A proper strategy might benefit both confidence-accuracy matching and confidence separability.
As shown in Section 4, calibration and failure prediction could be improved concurrently.
\setParDef

\section{Improving Failure Prediction by Finding Flat Minima}
As reported in Section 3, none of those popular calibration methods seem to address failure prediction better than simple baseline \cite{hendrycks2017baseline}. Does there exist a more principled and hassle-free strategy to improve failure prediction? 

\subsection{Motivation and Methodology}
\begin{figure*}[t]
	\begin{center}
		%\vskip -0.1 in
		\centerline{\includegraphics[width=1.05\textwidth]{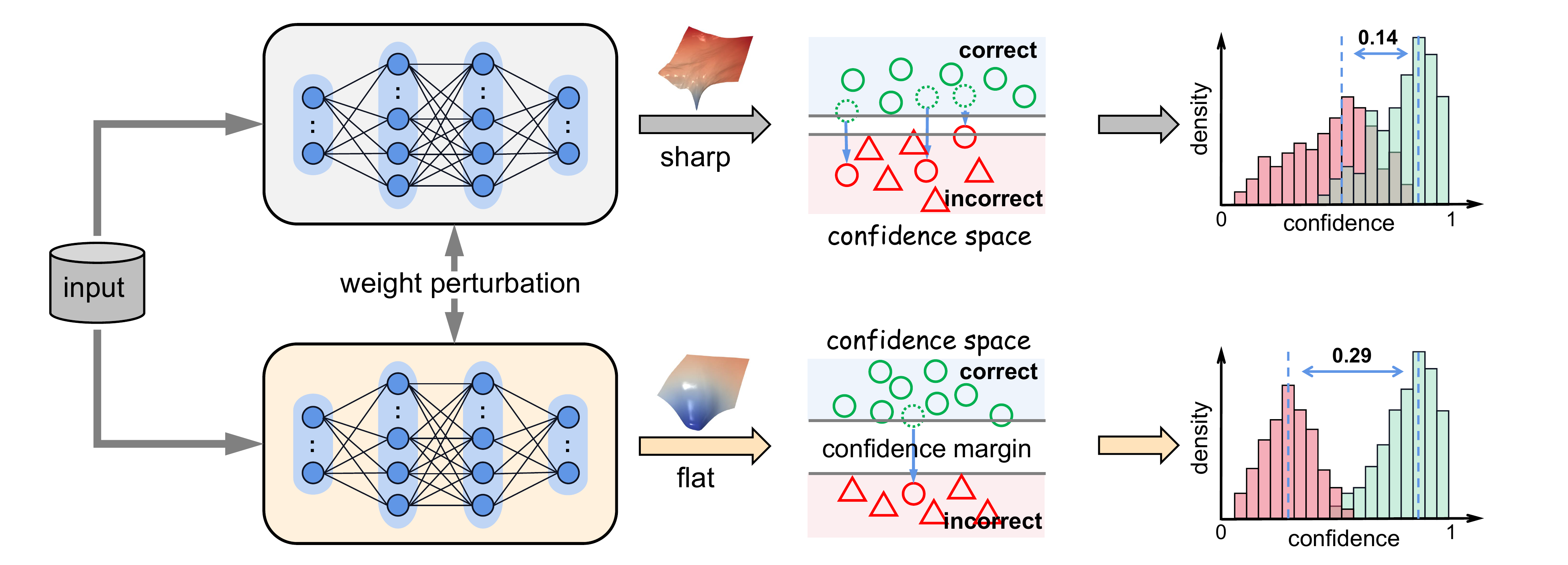}}
		\vskip -0.1 in
		\caption{An intuitive relationship between flatness and failure prediction. The stability of predictions to parameter perturbations can be seen as a \textbf{\emph{confidence margin}} condition. This inspires us to improve failure prediction by finding flat minima.}
		\label{figure-8}
	\end{center}
	\vskip -0.6 in
\end{figure*}

\noindent
\textbf{Rationale: why.} Confidence separability between correct and incorrect samples is crucial for failure prediction. Let us consider how confidence separability affects the confidence robustness of correct samples. Specifically, for a correctly classified sample, to become misclassified, it must reduce the probability on the ground truth class and increase its probability on another (wrong) class. During this process, the confidence margin plays a crucial role: a larger confidence margin could make it harder to change the predicted class label. 
Interestingly, flatness of a model reflects how sensitive the correct samples become misclassified when perturbing the weights of a model \cite{huang2020understanding, izmailov2018averaging, foret2020sharpness}. As illustrated in Fig.~\ref{figure-8}, with flat minima, a correct sample is difficult to be misclassified under weight perturbations and vice versa. Therefore, we conjecture that the confidence gap
between correct and incorrect samples of a flat minima is larger than that of a sharp minima.

\begin{wrapfigure}{r}{6cm}
	\centering
	\vskip -0.4 in
	\includegraphics[width=6cm]{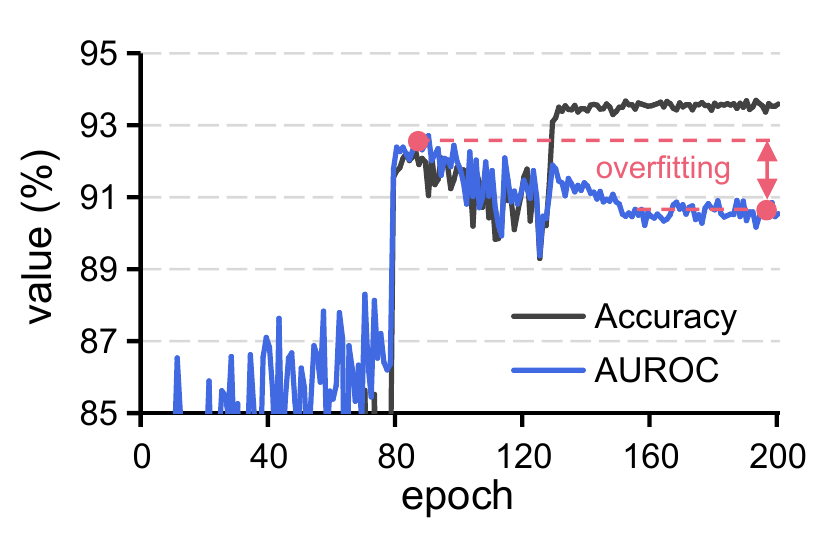}
	\vskip -0.2 in
	\caption{\emph{Reliable overfitting} phenomenon.}
	\label{figure-9}
	\vskip -0.25 in
\end{wrapfigure} 
\setParDis
\noindent
\textbf{Reliable overfitting phenomenon.} As shown in Fig.~\ref{figure-9}, we observed an interesting phenomenon that the AUROC can be easily overfitting during the training of a model (ResNet110 on CIFAR-10). Concretely, the test accuracy continually increases while the AUROC decreases at the last phases, making it difficult for failure prediction.
We term this phenomenon as ``\textbf{\emph{reliable overfitting}}'', which exists on different model and dataset settings and somewhat similar to the \emph{robust overfitting} \cite{rice2020overfitting} in adversarial robustness literature. Since flat minima has been verified to be effective for alleviating robust overfitting \cite{WuX020, chen2020robust}, we expect that flat minima could also benefit failure prediction.

\begin{table}[!t]
	\caption{Confidence estimation results. AURC and E-AURC values are multiplied by $10^3$, and NLL are multiplied by 10 for clarity. Remaining values are percentages.}
	\vskip -0.2in
	\label{table-2}
	\begin{center}
		\renewcommand\tabcolsep{4.2pt}
		\begin{small}
			\newcommand{\tabincell}[2]{\begin{tabular}{@{}#1@{}}#2\end{tabular}}
			\scalebox{0.58}{
				\renewcommand{\arraystretch}{1}
				\begin{tabular}{llccccccccc}
					\toprule
					\multicolumn{11}{c}{\textbf{CIFAR-10}}  \\ \midrule
					\textbf{Network} 
					& \textbf{Method}&
					\tabincell{c}{\textbf{AURC} \\ \textbf{($\downarrow$)}} & \tabincell{c}{\textbf{E-AURC} \\ \textbf{($\downarrow$)}}&
					\tabincell{c}{\textbf{FPR-95\%} \\ \textbf{TPR($\downarrow$)}} &
					\tabincell{c}{\textbf{AUROC} \\ \textbf{($\uparrow$)}} & \tabincell{c}{\textbf{AUPR-} \\ \textbf{Success($\uparrow$)}} & \tabincell{c}{\textbf{AUPR-} \\ \textbf{Error($\uparrow$)}} &
					\tabincell{c}{\textbf{ECE} \\ \textbf{($\downarrow$)}} & \tabincell{c}{\textbf{NLL} \\ \textbf{($\downarrow$)}}&
					\tabincell{c}{\textbf{Brier} \\ \textbf{($\downarrow$)}}\\
					\midrule
					\multirow{5}{*}{ResNet110}
					& baseline \cite{hendrycks2017baseline} & 9.94$\pm$1.29 & 7.94$\pm$1.23 & 45.01$\pm$2.55 & 90.76$\pm$0.60 & 99.16$\pm$0.13 & 43.66$\pm$1.86 & 4.14$\pm$0.06 & 2.88$\pm$0.08 & 10.47$\pm$0.28 \\
					& CRL \cite{MoonKSH20} & 7.54$\pm$0.20 & 5.29$\pm$0.16 & 45.25$\pm$3.24 & 92.86$\pm$0.17 & 99.44$\pm$0.02 & 44.93$\pm$1.31 & 1.65$\pm$0.08 & 2.05$\pm$0.03 & 10.01$\pm$0.09 \\ 
					\cmidrule{2-11} 
					& SAM & \bftab{5.31$\pm$0.13} & \bftab{3.74$\pm$0.17} & 39.44$\pm$2.75 & 93.73$\pm$0.45 & 99.61$\pm$0.02 & 45.23$\pm$2.88 & 1.86$\pm$0.12 & 1.76$\pm$0.02 & 8.55$\pm$0.19 \\
					& SWA & 6.38$\pm$0.18 & 4.38$\pm$0.07 & \bftab{39.36$\pm$0.29} & 93.69$\pm$0.08 & 99.54$\pm$0.01 & \bftab{47.49$\pm$2.03} & 1.33$\pm$0.10 & 1.83$\pm$0.03 & 9.17$\pm$0.15 \\
					& ours & 5.57$\pm$0.11 & 3.92$\pm$0.09 & 39.50$\pm$0.52 & \bftab{93.76$\pm$0.10} & \bftab{99.58$\pm$0.01} & 44.45$\pm$1.89 & \bftab{0.50$\pm$0.13} & \bftab{1.68$\pm$0.02} & \bftab{8.50$\pm$0.13} \\

					\midrule
					\multirow{5}{*}{WRNet}
					& baseline \cite{hendrycks2017baseline} & 4.89$\pm$0.25 & 3.93$\pm$0.22 & 32.85$\pm$0.36 & 93.24$\pm$0.15 & 99.59$\pm$0.02 & 43.38$\pm$1.12 & 2.71$\pm$0.06 & 1.82$\pm$0.03 & 7.15$\pm$0.07 \\
					& CRL \cite{MoonKSH20} & 4.52$\pm$0.10 & 3.23$\pm$0.10 & 32.92$\pm$2.18 & 94.24$\pm$0.23 & 99.66$\pm$0.01 & 44.07$\pm$1.61 & 0.44$\pm$0.14 & 1.52$\pm$0.02 & 7.41$\pm$0.14 \\ 
					\cmidrule{2-11} 
					& SAM & 2.85$\pm$0.04 & 2.15$\pm$0.07 & \bftab{26.07$\pm$0.96} & 95.10$\pm$0.19 & 99.78$\pm$0.01 & 44.50$\pm$1.20 & 1.58$\pm$0.05 & 1.25$\pm$0.01 & 5.76$\pm$0.07 \\
					& SWA & 2.84$\pm$0.04 & 2.06$\pm$0.05 & 28.60$\pm$0.08 & 95.31$\pm$0.12 & 99.79$\pm$0.01 & \bftab{44.62$\pm$2.14} & 1.22$\pm$0.05 & 1.22$\pm$0.01 & 5.92$\pm$0.05 \\
					& ours & \bftab{2.60$\pm$0.06} & \bftab{1.90$\pm$0.03} & 27.80$\pm$2.19 & \bftab{95.43$\pm$0.07} & \bftab{99.80$\pm$0.01} & 44.36$\pm$3.46& \bftab{0.43$\pm$0.14} & \bftab{1.12$\pm$0.01} & \bftab{5.57$\pm$0.03} \\
					
					\midrule
					\multirow{5}{*}{DenseNet}
					& baseline \cite{hendrycks2017baseline} & 6.20$\pm$0.29 & 4.66$\pm$0.27 & 38.20$\pm$2.48  & 92.87$\pm$0.44 & 99.51$\pm$0.03 & 43.74$\pm$2.44 & 2.96$\pm$0.12 & 2.10$\pm$0.06 & 8.77$\pm$0.17 \\
					& CRL \cite{MoonKSH20} & 6.17$\pm$0.26 & 4.29$\pm$0.09 & 39.76$\pm$0.24 & 93.64$\pm$0.18 & 99.55$\pm$0.01 & 46.51$\pm$1.88 & 0.95$\pm$0.18 & 1.81$\pm$0.06 & 8.98$\pm$0.35 \\ 
					\cmidrule{2-11} 
					& SAM & 4.56$\pm$0.17 & 3.06$\pm$0.14 & 33.69$\pm$1.68 & 94.55$\pm$0.20 & 99.68$\pm$0.01 & 47.32$\pm$1.03 & 1.37$\pm$0.13 & 1.53$\pm$0.05 & \bftab{7.45$\pm$0.20} \\
					& SWA & 5.02$\pm$0.20 & 3.48$\pm$0.04 & 37.87$\pm$2.71 & 94.31$\pm$0.24 & 99.64$\pm$0.01 & 44.48$\pm$1.38 & 1.13$\pm$0.11 & 1.61$\pm$0.03 & 8.12$\pm$0.24 \\
					& ours & \bftab{4.48$\pm$0.13} & \bftab{3.01$\pm$0.12} & \bftab{31.41$\pm$1.29} & \bftab{94.92$\pm$0.19} & \bftab{99.69$\pm$0.01} & \bftab{48.12$\pm$0.88} & \bftab{0.52$\pm$0.08} & \bftab{1.51$\pm$0.01} & 7.68$\pm$0.06 \\

					\midrule
					\multirow{5}{*}{ConvMixer}
					& baseline \cite{hendrycks2017baseline} & 8.33$\pm$1.44 & 6.29$\pm$1.30 & 42.32$\pm$3.26 & 92.02$\pm$0.96 & 99.34$\pm$0.14 & 43.80$\pm$1.49 & 3.43$\pm$0.51 & 2.53$\pm$0.27 & 10.15$\pm$0.65 \\
					& CRL \cite{MoonKSH20} & 6.89$\pm$0.51 & 4.78$\pm$0.35 & 41.47$\pm$1.38 & 93.27$\pm$0.21 & 99.50$\pm$0.04 & 45.99$\pm$1.38 & 0.99$\pm$0.22 & 1.98$\pm$0.07 & 9.57$\pm$0.31 \\ 
					\cmidrule{2-11} 
					& SAM & 5.52$\pm$0.22 & 3.88$\pm$0.16 & 36.16$\pm$1.35 & 93.92$\pm$0.32 & 99.59$\pm$0.02 & 46.66$\pm$2.60 & 2.16$\pm$0.21 & 1.84$\pm$0.06 & 8.55$\pm$0.29 \\
					& SWA & \bftab{4.68$\pm$0.26} & \bftab{3.35$\pm$0.17} & 34.73$\pm$2.04 & 94.54$\pm$0.16 & 99.65$\pm$0.02 & 45.62$\pm$1.17 & 1.31$\pm$0.17 & 1.59$\pm$0.07 & 7.90$\pm$0.30\\
					& ours & 4.98$\pm$0.23 & 3.47$\pm$0.13 & \bftab{32.88$\pm$1.77} & \bftab{94.75$\pm$0.19} & \bftab{99.63$\pm$0.02} & \bftab{49.02$\pm$1.36} & \bftab{0.89$\pm$0.20} & \bftab{1.58$\pm$0.05} & \bftab{7.86$\pm$0.28} \\

					\bottomrule
					\toprule
					\multicolumn{11}{c}{\textbf{CIFAR-100}}  \\ \midrule
					\textbf{Network} 
					& \textbf{Method}&
					\tabincell{c}{\textbf{AURC} \\ \textbf{($\downarrow$)}} & \tabincell{c}{\textbf{E-AURC} \\ \textbf{($\downarrow$)}}&
					\tabincell{c}{\textbf{FPR-95\%} \\ \textbf{TPR($\downarrow$)}} &
					\tabincell{c}{\textbf{AUROC} \\ \textbf{($\uparrow$)}} & \tabincell{c}{\textbf{AUPR-} \\ \textbf{Success($\uparrow$)}} & \tabincell{c}{\textbf{AUPR-} \\ \textbf{Error($\uparrow$)}} &
					\tabincell{c}{\textbf{ECE} \\ \textbf{($\downarrow$)}} & \tabincell{c}{\textbf{NLL} \\ \textbf{($\downarrow$)}}&
					\tabincell{c}{\textbf{Brier} \\ \textbf{($\downarrow$)}}\\
					\midrule
					\multirow{5}{*}{ResNet110}
					& baseline \cite{hendrycks2017baseline} & 93.90$\pm$2.37 & 50.88$\pm$2.03 & 66.02$\pm$1.53 & 85.00$\pm$0.35 & 93.42$\pm$0.28 & 66.34$\pm$0.29 & 15.71$\pm$0.15 & 13.54$\pm$0.12 & 42.88$\pm$0.27 \\
					& CRL \cite{MoonKSH20} & 81.02$\pm$2.06 & 41.82$\pm$0.94 & 64.37$\pm$0.43 & 86.32$\pm$0.24 & 94.68$\pm$0.13 & 66.26$\pm$0.32 & 10.96$\pm$0.53 & 11.02$\pm$0.09 & 38.82$\pm$0.57 \\ 
					\cmidrule{2-11} 
					& SAM & 78.35$\pm$1.32 & 41.01$\pm$0.61 & 63.86$\pm$0.52 & 86.38$\pm$0.14 & 94.82$\pm$0.09 & 66.21$\pm$0.72 & 10.58$\pm$0.15 & 10.54$\pm$0.18 & 37.89$\pm$0.44 \\
					& SWA & 71.94$\pm$0.60 & 37.23$\pm$0.67 & 63.98$\pm$0.68 & 86.86$\pm$0.21 & 95.35$\pm$0.08 & 65.59$\pm$0.88 & 5.38$\pm$0.13 & 8.81$\pm$0.02 & 34.95$\pm$0.08 \\
					& ours & \bftab{69.51$\pm$0.56} & \bftab{35.51$\pm$0.08} & \bftab{62.57$\pm$0.81} & \bftab{87.24$\pm$0.12} & \bftab{95.57$\pm$0.02} & \bftab{66.52$\pm$1.04} & \bftab{3.41$\pm$0.35} & \bftab{8.51$\pm$0.07} & \bftab{34.10$\pm$0.19} \\

					\midrule
					\multirow{5}{*}{WRNet}
					& baseline \cite{hendrycks2017baseline} & 51.97$\pm$1.74 & 30.18$\pm$1.09 & 59.19$\pm$0.20 & 87.75$\pm$0.30 & 96.38$\pm$0.14 & 62.99$\pm$0.98 & 7.19$\pm$0.36 & 8.43$\pm$0.11 & 29.67$\pm$0.45 \\
					& CRL \cite{MoonKSH20} & 46.17$\pm$0.20 & 25.28$\pm$0.34 & 58.32$\pm$0.48 & 88.77$\pm$0.35 & 97.00$\pm$0.03 & \bftab{63.34$\pm$0.98} & 3.93$\pm$0.06 & 7.33$\pm$0.06 & 27.98$\pm$0.17 \\ 
					\cmidrule{2-11} 
					& SAM & 43.21$\pm$0.22 & 24.51$\pm$0.65 & 57.56$\pm$0.91 & 88.82$\pm$0.26 & 97.11$\pm$0.07 & 62.69$\pm$0.72 & 4.97$\pm$0.38 & 7.15$\pm$0.03 & 26.99$\pm$0.16 \\
					& SWA & 41.62$\pm$0.18 & 22.73$\pm$0.02 & 57.19$\pm$0.87 & 89.23$\pm$0.03 & 97.33$\pm$0.01 & 62.17$\pm$0.27 & 7.61$\pm$0.14 & 7.29$\pm$0.05 & 27.64$\pm$0.06 \\
					& ours & \bftab{40.80$\pm$0.31} & \bftab{22.00$\pm$0.15} & \bftab{56.13$\pm$0.44} & \bftab{89.53$\pm$0.10} & \bftab{97.41$\pm$0.01} & 63.11$\pm$0.12 & \bftab{6.07$\pm$0.34} & \bftab{6.83$\pm$0.01} & \bftab{26.78$\pm$0.20} \\
					\midrule

					\multirow{5}{*}{DenseNet}
					& baseline \cite{hendrycks2017baseline} & 67.41$\pm$0.67 & 35.82$\pm$0.54 & 61.55$\pm$2.01 & 86.46$\pm$0.31 & 95.55$\pm$0.05 & \bftab{65.60$\pm$1.26} & 9.04$\pm$0.31 & 9.60$\pm$0.08 & 35.11$\pm$0.39 \\
					& CRL \cite{MoonKSH20} & 64.30$\pm$1.26 & 34.59$\pm$0.33 & 61.42$\pm$1.74 & 87.19$\pm$0.25 & 95.74$\pm$0.06 & 64.67$\pm$1.45 & 5.38$\pm$0.34 & 8.50$\pm$0.14 & 32.85$\pm$0.32 \\ 
					\cmidrule{2-11} 
					& SAM & 63.52$\pm$2.41 & 34.51$\pm$1.07 & 61.46$\pm$0.32  & 87.01$\pm$0.13 & 95.76$\pm$0.15 & 64.41$\pm$1.04 & 5.91$\pm$0.39 & 8.53$\pm$0.18 & 32.75$\pm$0.71 \\
					& SWA & 59.88$\pm$1.40 & 32.08$\pm$0.25 & 63.11$\pm$0.73 & 87.34$\pm$0.18 & 96.09$\pm$0.05 & 62.88$\pm$1.39 & 4.89$\pm$0.28 & 7.82$\pm$0.10 & 31.63$\pm$0.47 \\
					& ours & \bftab{56.62$\pm$0.18} & \bftab{30.33$\pm$0.42} & \bftab{61.34$\pm$1.59} & \bftab{87.75$\pm$0.20} & \bftab{96.32$\pm$0.05} & 63.45$\pm$0.70 & \bftab{3.16$\pm$0.08} & \bftab{7.53$\pm$0.04} & \bftab{30.55$\pm$0.10} \\

					\midrule
					\multirow{5}{*}{ConvMixer}
					& baseline \cite{hendrycks2017baseline} & 76.96$\pm$1.64 & 41.20$\pm$1.64 & 63.57$\pm$0.52 & 86.28$\pm$0.18 & 94.81$\pm$0.06 & 65.39$\pm$0.83 & 7.42$\pm$1.20 & 9.98$\pm$0.33 & 36.27$\pm$0.83 \\
					& CRL \cite{MoonKSH20} & 64.20$\pm$0.50 & 33.57$\pm$0.56 & 60.11$\pm$2.62 & 87.59$\pm$0.31 & 95.85$\pm$0.06 & \bftab{65.78$\pm$1.73} & 3.99$\pm$0.43 & 8.55$\pm$0.02 & 32.84$\pm$0.06 \\ 
					\cmidrule{2-11}
					& SAM & 64.47$\pm$2.10 & 34.05$\pm$1.19 & 64.58$\pm$1.44 & 87.11$\pm$0.24 & 95.82$\pm$0.16 & 63.52$\pm$0.66 & 7.96$\pm$0.10 & 8.34$\pm$0.10 & 33.36$\pm$0.38 \\
					& SWA & 69.73$\pm$2.51 & 37.47$\pm$1.14 & 63.43$\pm$1.24 & 86.56$\pm$0.15 & 95.35$\pm$0.17 & 64.57$\pm$0.64  & \bftab{6.29$\pm$0.35} & 9.13$\pm$0.24 & 34.54$\pm$0.67 \\
					& ours & \bftab{57.15$\pm$1.92} & \bftab{29.96$\pm$0.59} & \bftab{60.49$\pm$2.22} & \bftab{88.04$\pm$0.35} & \bftab{96.35$\pm$0.09} & 64.57$\pm$2.05 & 6.40$\pm$0.62 & \bftab{8.05$\pm$0.18} & \bftab{31.52$\pm$0.63} \\
					
					\bottomrule
			\end{tabular}}
		\end{small}
	\end{center}
	\vskip -0.25in
\end{table}

\noindent
\textbf{Approach: how.} There are several methods have been proposed to seek flat minima for DNNs \cite{izmailov2018averaging, foret2020sharpness, pittorino2021entropic, chaudhari2019entropy}. We select \emph{stochastic weight averaging} (SWA) \cite{izmailov2018averaging} and \emph{sharpness-aware minimization} (SAM) \cite{foret2020sharpness} as two representative methods due to the simplicity for proofs-of-concept. Specifically, SWA simply averages multiple parameters of the model along the training trajectory: $\bm{\theta}_{swa}^t = \frac{{\bm{\theta}_{swa}}^{t-1} \times n + \bm{\theta}^t}{n+1}$,
where $n$ indexs the number of past checkpoints to be averaged, $t$ is the training epoch, $\bm{\theta}$ is the current weights, $\bm{\theta}_{swa}$ is the averaged weights. While SAM finds the flat minima by directly perturbing the weights: $\min\limits_{\bm{\theta}}\max\limits_{||\epsilon||_p \le \rho} \mathcal{L}(\bm{\theta} + \epsilon) + \frac{\lambda}{2}||\bm{\theta}||$. Although the SWA and SAM find flat minima based on different mechanism, we find  they both improve the failure prediction performance. This also motivates us to combine them to get better performance. We refer the combine of them as \textbf{FMFP} (\emph{\textbf{F}lat \textbf{M}inima for \textbf{F}ailure \textbf{P}rediction}). Supplementary material presents pseudo-code for the FMFP algorithm.
\setParDef

\begin{figure*}[t]
	\begin{center}
		\centerline{\includegraphics[width=1.04\textwidth]{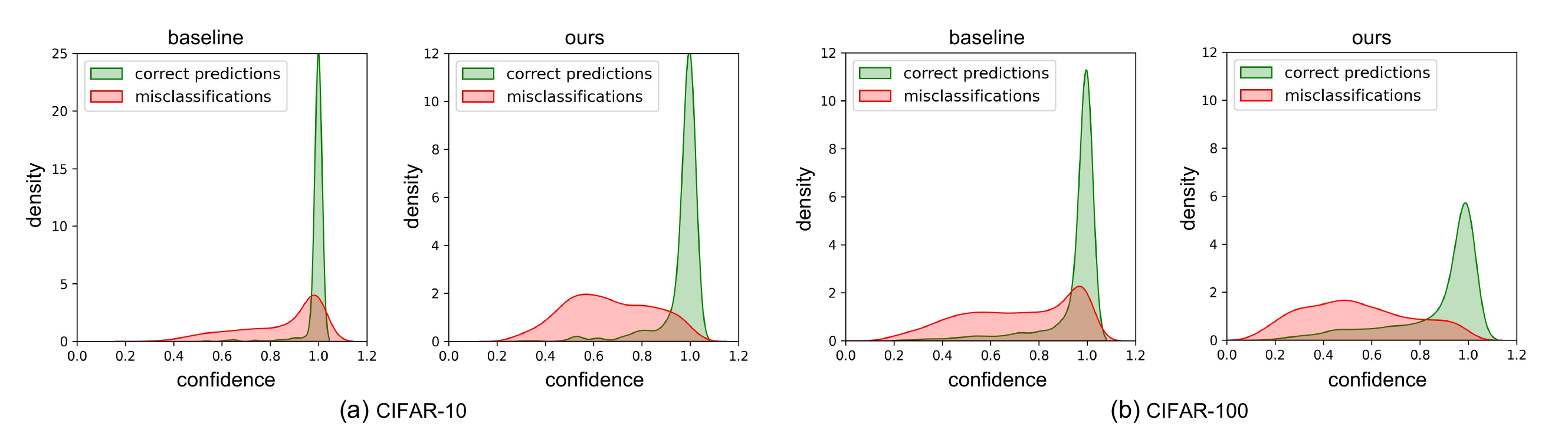}}
		\vskip -0.17 in
		\caption{Confidence distribution on correct and misclassified samples. Our method leads to a better separation for failure prediction.}
		\label{figure-10}
	\end{center}
	\vskip -0.35 in
\end{figure*}
\begin{figure*}[t]
	\begin{center}
		\centerline{\includegraphics[width=0.9\textwidth]{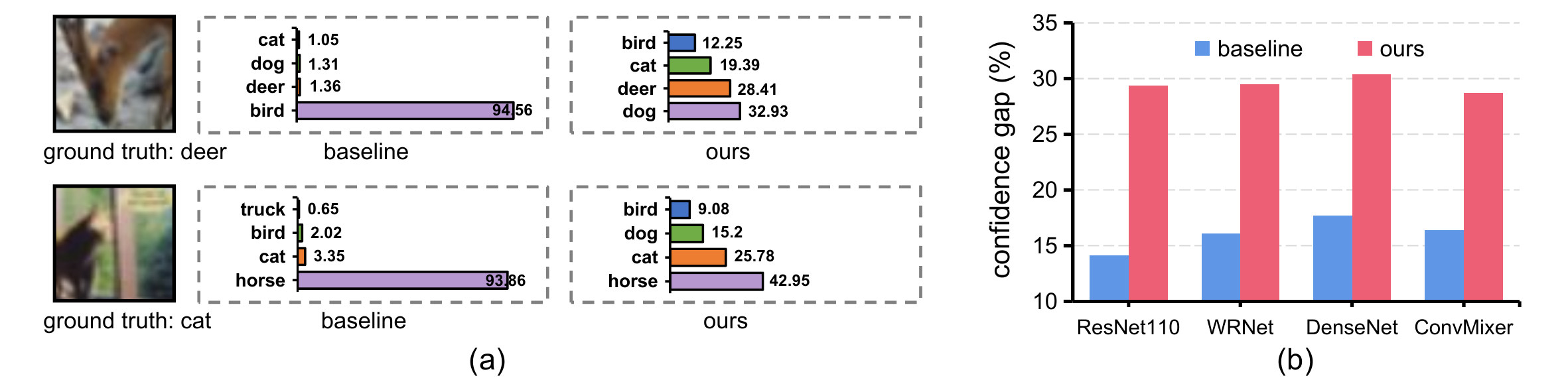}}
		\vskip -0.15 in
		\caption{(a) Predictive distribution on misclassified samples. (b) Our method (FMFP) significantly enlarges the average confidence gap between correct and incorrect samples.}
		\label{figure-11}
	\end{center}
	\vskip -0.43 in
\end{figure*}

\subsection{Experiments Results and Analysis}
\textbf{Experimental setup.} We conduct experiments on CIFAR-10, CIFAR-100, and Tiny-ImageNet with various network architectures. For comparison methods, we mainly compare our method with baseline \cite{hendrycks2017baseline} and CRL \cite{MoonKSH20}, which is the state-of-the-art approach for failure prediction that outperforms representative bayesian methods \cite{GalG16, KendallG17}. Due to the limitation of space, we provide implementation details and experimental results on Tiny-ImageNet in supplementary material.

\setParDis
\noindent
\textbf{Flat minima do improve failure prediction.} Comparative results are summarized in Table~\ref{table-2}. We observe that flat minima based methods: SAM, SWA, and FMFP (ours) consistently outperform the strong baseline and CRL on various metrics of failure prediction. Particularly, FMFP generally yields the best results. 
For example, in the case of ResNet110, our method has 3.00\% and 2.24\% higher values of AUROC on CIFAR-10 and CIFAR-100, respectively. In addition, flat minima based methods, especially the proposed FMFP, can achieve effective gains over confidence calibration (the last three columns in Table~\ref{table-2}).

\noindent
In Fig.~\ref{figure-10}, we observe that correct predictions and erroneous predictions overlap severely, making it difficult to distinguish them. Our method remarkably shifts the errors’ confidence distributions to smaller values and maintains the
\begin{wrapfigure}{r}{7cm}
	\centering
	\vskip -0.3 in
	\includegraphics[width=7cm]{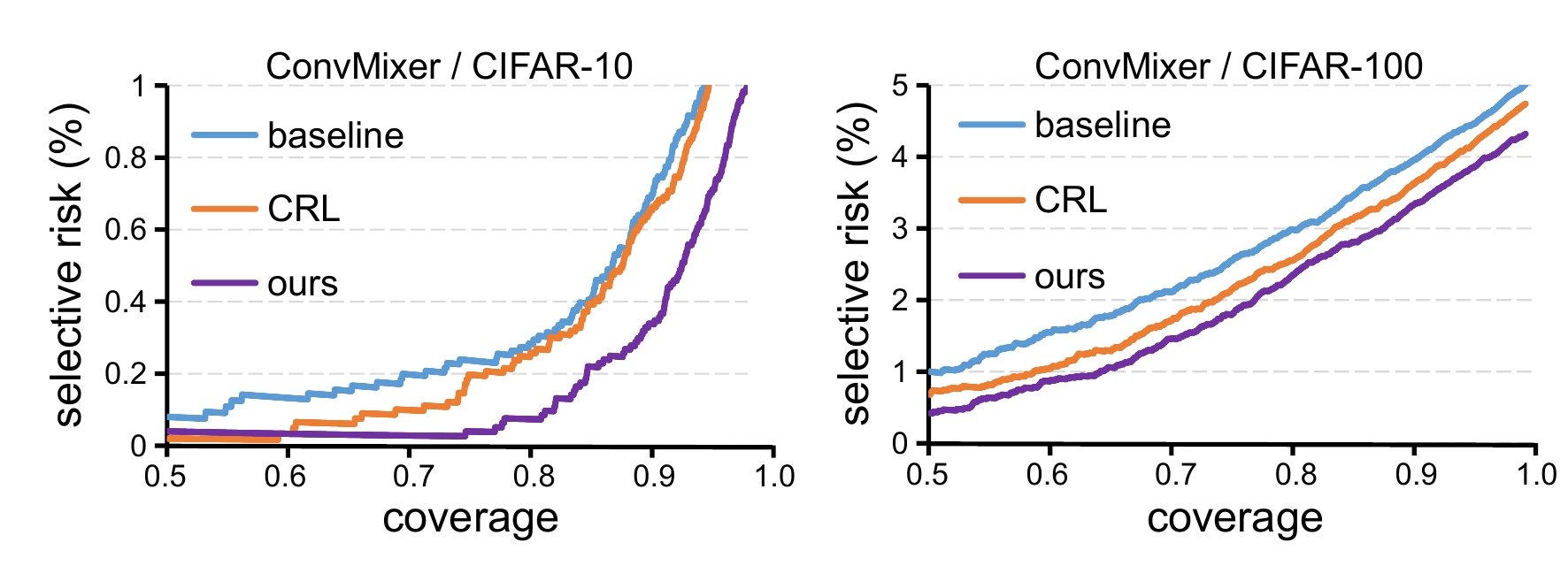}
	\vskip -0.17 in
	\caption{Comparison of risk-coverage curves.}
	\label{figure-12}
	\vskip -0.2 in
\end{wrapfigure}
 confidence of correct samples, leading to a better separation
for failure prediction. Fig.~\ref{figure-11}(a) presents some examples of misclassified samples and their corresponding confidence distribution.
Ours outputs much lower confidence on the erroneously predicted class. In Fig.~\ref{figure-11}(b), our method significantly enlarges the confidence gap between correct and incorrect samples. Besides, the risk-coverage curves in Fig.~\ref{figure-12} and Fig.~\ref{figure-2}(b) also demonstrate the confidence reliability of our method.

\noindent
\textbf{Failure prediction under distribution shift.} In real-world applications, the model may encoder inputs subject to various kinds of distributional shifts. Thus, it becomes necessary to evaluate the confidence estimation performance to distributional shifts. The model is trained on CIFAR-10 and evaluated on corrupted dataset CIFAR-10-C \cite{HendrycksD19}. The average results for 15 kinds of corruption under 5 different levels of perturbation severity are reported in Table~\ref{table-3}. Our method consistently performs better than baseline and CRL methods.
\begin{table}[t]
	\caption{Failure prediction performance under distributional shifts.}
	\vskip -0.07in
	\label{table-3}
	\setlength\tabcolsep{2pt}
	\centering
	\renewcommand{\arraystretch}{1.1}
	\scalebox{0.6}{
		\begin{tabular}{lcccccccccccc}
			\toprule
			\multirow{2}{*}{\textbf{Method}} & \multicolumn{4}{c}{\textbf{AUROC} $\uparrow$} & \multicolumn{4}{c}{\textbf{AURC} $\downarrow$} & \multicolumn{4}{c}{\textbf{FPR-95\%TPR} $\downarrow$}\\
			\cmidrule(lr){2-5} \cmidrule(lr){6-9} \cmidrule(lr){10-13}
			& ResNet110 & WRNet & DenseNet & ConvMixer & ResNet110 & WRNet & DenseNet & ConvMixer & ResNet110 & WRNet & DenseNet & ConvMixer\\
			\midrule
			baseline \cite{hendrycks2017baseline} & 79.45 &83.81  &81.97  &81.28  &157.46  &112.46  &148.91  &168.39  &71.29  &64.05  & 69.26 & 71.07  \\
			CRL \cite{MoonKSH20} & 82.54 &85.91  &83.51  &82.46  &133.73  &104.95  &133.58  &163.68  &68.86  &63.35  &67.73  &69.34  \\
			ours & \bftab{84.72} &\bftab{87.34}  &\bftab{84.90}  &\bftab{84.93} &\bftab{119.79}  &\bftab{94.34}  &\bftab{130.42}  &\bftab{145.21}  &\bftab{65.51}  &\bftab{58.87}  &\bftab{64.95}  &\bftab{65.29} \\
			\bottomrule
		\end{tabular}
	}
	\vskip -0.15in
\end{table}

\begin{wrapfigure}{r}{6cm}
	\centering
	\vskip -0.07 in
	\includegraphics[width=6cm]{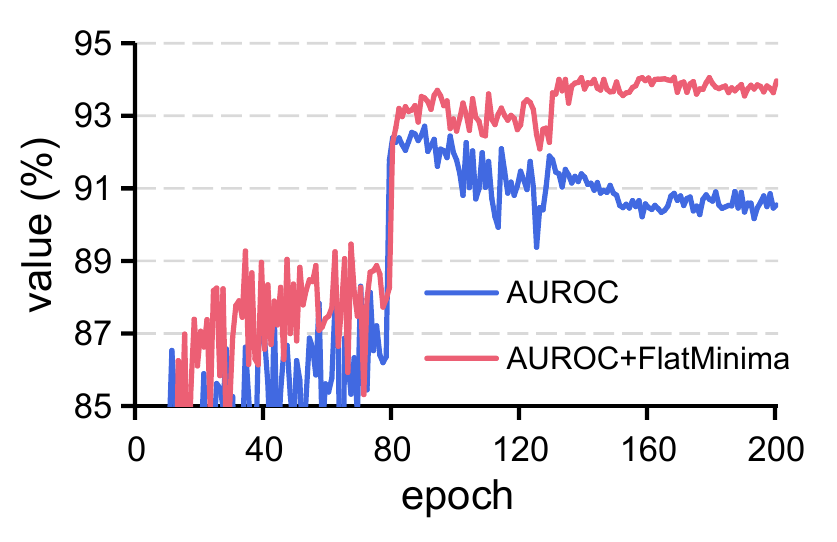}
	\vskip -0.2 in
	\caption{Flat minima effectively mitigates the reliable
		overfitting.}
	\label{figure-13}
	\vskip -0.25 in
\end{wrapfigure} 
\noindent
\textbf{Flat minima mitigates the reliable overfitting.} Fig.~\ref{figure-13} plots the AUROC curves during training. 
Although the failure prediction performance can be improved by early stopping, the classification accuracy of early checkpoint is much lower (Fig.~\ref{figure-9}). We can clearly observe that with flat minima, reliable overfitting has been diminished significantly, and the AUROC curves robustly improve until the end. Flat minima further leads to better classification accuracy, avoiding the trade-off between AUROC and accuracy when applying early stopping.

\noindent
\textbf{Further understanding flat minima for failure prediction.}
\ding{172} \textbf{\emph{Representation learning.}}  Misclassification with high confidence is often attributed to spurious correlations appearing in the sample and the wrong class. Flat minima has been theoretically proved to result in invariant and disentangled representations \cite{AchilleS18}, which is effective for spurious representations mitigation \cite{cha2021swad}. Therefore, with fewer spurious or irrelevant representations, the misclassified sample would be near the decision boundary with low confidence and less activated for wrong classes.
\ding{173} \textbf{\emph{Uncertainty.}} It has been shown that flat minima corresponds to regions in parameter space with rich posterior uncertainty \cite{pml1Book}. Therefore, flat minima has the advantage to indicate the uncertainty of an input.

\setParDef
\section{Concluding Remarks}
Failure prediction is an important yet far less explored problem for safety-critical applications. This paper evaluates the effect of popular calibration methods for failure prediction. To our surprise, they have no or negative effect on failure prediction. We further find the under-confident issue of correctly classified samples, which leads to worse separation between the confidence of correct and incorrect samples. Finally, we propose to enlarge the confidence gap by finding flat minima, which yields strong performance on extensive experiments.

\setParDis
\noindent
\textbf{Acknowledgement.}
This work has been supported by the National Key Research and Development Program under Grant No. 2018AAA0100400, the National Natural Science Foundation of China grants U20A20223, 62076236, 61721004, the Key Research Program of Frontier Sciences of CAS under Grant ZDBS-LY-7004, and the Youth Innovation Promotion Association of CAS under Grant 2019141.
\setParDef

\clearpage
\bibliographystyle{splncs04}
\bibliography{reference}

\begin{thebibliography}{10}
\providecommand{\url}[1]{\texttt{#1}}
\providecommand{\urlprefix}{URL }
\providecommand{\doi}[1]{https://doi.org/#1}

\bibitem{AchilleS18}
Achille, A., Soatto, S.: Emergence of invariance and disentanglement in deep
  representations. J. Mach. Learn. Res.  \textbf{19},  50:1--50:34 (2018)

\bibitem{bojarski2016end}
Bojarski, M., Del~Testa, D., Dworakowski, D., Firner, B., Flepp, B., Goyal, P.,
  Jackel, L.D., Monfort, M., Muller, U., Zhang, J., et~al.: End to end learning
  for self-driving cars. arXiv preprint arXiv:1604.07316  (2016)

\bibitem{Brier1950}
Brier, G.W.: Verification of forecasts expressed in terms of probability.
  Monthly Weather Review pp.~1--3 (1950)

\bibitem{brier1950verification}
Brier, G.W., et~al.: Verification of forecasts expressed in terms of
  probability. Monthly weather review  \textbf{78}(1), ~1--3 (1950)

\bibitem{cha2021swad}
Cha, J., Chun, S., Lee, K., Cho, H.C., Park, S., Lee, Y., Park, S.: Swad:
  Domain generalization by seeking flat minima. In: NeurIPS (2021)

\bibitem{chaudhari2019entropy}
Chaudhari, P., Choromanska, A., Soatto, S., LeCun, Y., Baldassi, C., Borgs, C.,
  Chayes, J., Sagun, L., Zecchina, R.: Entropy-sgd: Biasing gradient descent
  into wide valleys. Journal of Statistical Mechanics: Theory and Experiment
  \textbf{2019}(12),  124018 (2019)

\bibitem{chen2020robust}
Chen, T., Zhang, Z., Liu, S., Chang, S., Wang, Z.: Robust overfitting may be
  mitigated by properly learned smoothening. In: ICLR (2021)

\bibitem{corbiere2019addressing}
Corbi{\`e}re, C., Thome, N., Bar-Hen, A., Cord, M., P{\'e}rez, P.: Addressing
  failure prediction by learning model confidence. In: NeurIPS. pp. 2898--2909
  (2019)

\bibitem{corbiere2021confidence}
Corbi{\`e}re, C., Thome, N., Saporta, A., Vu, T.H., Cord, M., Perez, P.:
  Confidence estimation via auxiliary models. IEEE Transactions on Pattern
  Analysis and Machine Intelligence  (2021)

\bibitem{deng2009imagenet}
Deng, J., Dong, W., Socher, R., Li, L.J., Li, K., Fei-Fei, L.: Imagenet: A
  large-scale hierarchical image database. In: CVPR. pp. 248--255 (2009)

\bibitem{dosovitskiy2020image}
Dosovitskiy, A., Beyer, L., Kolesnikov, A., Weissenborn, D., Zhai, X.,
  Unterthiner, T., Dehghani, M., Minderer, M., Heigold, G., Gelly, S., et~al.:
  An image is worth 16x16 words: Transformers for image recognition at scale.
  In: ICLR (2020)

\bibitem{esteva2017dermatologist}
Esteva, A., Kuprel, B., Novoa, R.A., Ko, J., Swetter, S.M., Blau, H.M., Thrun,
  S.: Dermatologist-level classification of skin cancer with deep neural
  networks. nature  \textbf{542}(7639),  115--118 (2017)

\bibitem{foret2020sharpness}
Foret, P., Kleiner, A., Mobahi, H., Neyshabur, B.: Sharpness-aware minimization
  for efficiently improving generalization. In: ICLR (2020)

\bibitem{GalG16}
Gal, Y., Ghahramani, Z.: Dropout as a bayesian approximation: Representing
  model uncertainty in deep learning. In: ICML. pp. 1050--1059 (2016)

\bibitem{GeifmanE17}
Geifman, Y., El{-}Yaniv, R.: Selective classification for deep neural networks.
  In: NeurIPS. pp. 4878--4887 (2017)

\bibitem{GeifmanUE19}
Geifman, Y., Uziel, G., El{-}Yaniv, R.: Bias-reduced uncertainty estimation for
  deep neural classifiers. In: ICLR (2019)

\bibitem{guo2017calibration}
Guo, C., Pleiss, G., Sun, Y., Weinberger, K.Q.: On calibration of modern neural
  networks. In: ICML. pp. 1321--1330 (2017)

\bibitem{gupta2020calibration}
Gupta, K., Rahimi, A., Ajanthan, T., Mensink, T., Sminchisescu, C., Hartley,
  R.: Calibration of neural networks using splines. In: ICLR (2020)

\bibitem{havasi2020training}
Havasi, M., Jenatton, R., Fort, S., Liu, J.Z., Snoek, J., Lakshminarayanan, B.,
  Dai, A.M., Tran, D.: Training independent subnetworks for robust prediction.
  In: ICLR (2020)

\bibitem{he2016deep}
He, K., Zhang, X., Ren, S., Sun, J.: Deep residual learning for image
  recognition. In: CVPR. pp. 770--778 (2016)

\bibitem{HeZRS16}
He, K., Zhang, X., Ren, S., Sun, J.: Identity mappings in deep residual
  networks. In: ECCV. pp. 630--645 (2016)

\bibitem{Hebbalaguppe2022CVPR}
Hebbalaguppe, R., Prakash, J., Madan, N., Arora, C.: A stitch in time saves
  nine: A train-time regularizing loss for improved neural network calibration.
  In: CVPR. pp. 16081--16090 (June 2022)

\bibitem{HendrycksD19}
Hendrycks, D., Dietterich, T.G.: Benchmarking neural network robustness to
  common corruptions and perturbations. In: ICLR (2019)

\bibitem{hendrycks2017baseline}
Hendrycks, D., Gimpel, K.: A baseline for detecting misclassified and
  out-of-distribution examples in neural networks. In: ICLR (2017)

\bibitem{hendrycks2019deep}
Hendrycks, D., Mazeika, M., Dietterich, T.G.: Deep anomaly detection with
  outlier exposure. In: ICLR (2019)

\bibitem{HendrycksMCZGL20}
Hendrycks, D., Mu, N., Cubuk, E.D., Zoph, B., Gilmer, J., Lakshminarayanan, B.:
  Augmix: {A} simple data processing method to improve robustness and
  uncertainty. In: ICLR (2020)

\bibitem{howard2017mobilenets}
Howard, A.G., Zhu, M., Chen, B., Kalenichenko, D., Wang, W., Weyand, T.,
  Andreetto, M., Adam, H.: Mobilenets: Efficient convolutional neural networks
  for mobile vision applications. arXiv preprint arXiv:1704.04861  (2017)

\bibitem{HuangLMW17}
Huang, G., Liu, Z., van~der Maaten, L., Weinberger, K.Q.: Densely connected
  convolutional networks. In: CVPR. pp. 2261--2269 (2017)

\bibitem{huang2020understanding}
Huang, W.R., Emam, Z.A.S., Goldblum, M., Fowl, L.H., Terry, J., Huang, F.,
  Goldstein, T.: Understanding generalization through visualizations. In: ''I
  Can't Believe It's Not Better!''NeurIPS 2020 workshop (2020)

\bibitem{izmailov2018averaging}
Izmailov, P., Wilson, A., Podoprikhin, D., Vetrov, D., Garipov, T.: Averaging
  weights leads to wider optima and better generalization. In: UAI. pp.
  876--885 (2018)

\bibitem{janai2017computer}
Janai, J., G{\"u}ney, F., Behl, A., Geiger, A., et~al.: Computer vision for
  autonomous vehicles: Problems, datasets and state of the art. Foundations and
  Trends{\textregistered} in Computer Graphics and Vision  \textbf{12}(1--3),
  1--308 (2020)

\bibitem{Jiang2018ToTO}
Jiang, H., Kim, B., Gupta, M.R.: To trust or not to trust a classifier. In:
  NeurIPS (2018)

\bibitem{joo2020revisiting}
Joo, T., Chung, U.: Revisiting explicit regularization in neural networks for
  well-calibrated predictive uncertainty. arXiv preprint arXiv:2006.06399
  (2020)

\bibitem{KendallG17}
Kendall, A., Gal, Y.: What uncertainties do we need in bayesian deep learning
  for computer vision? In: NeurIPS. pp. 5574--5584 (2017)

\bibitem{krizhevsky2009learning}
Krizhevsky, A., Hinton, G., et~al.: Learning multiple layers of features from
  tiny images. Tech. rep., Citeseer (2009)

\bibitem{KullPKFSF19}
Kull, M., Perell{\'{o}}{-}Nieto, M., K{\"{a}}ngsepp, M., de~Menezes~e
  Silva~Filho, T., Song, H., Flach, P.A.: Beyond temperature scaling: Obtaining
  well-calibrated multi-class probabilities with dirichlet calibration. In:
  NeurIPS. pp. 12295--12305 (2019)

\bibitem{KullFF17}
Kull, M., de~Menezes~e Silva~Filho, T., Flach, P.A.: Beta calibration: a
  well-founded and easily implemented improvement on logistic calibration for
  binary classifiers. In: AISTATS. pp. 623--631 (2017)

\bibitem{kumar2019verified}
Kumar, A., Liang, P.S., Ma, T.: Verified uncertainty calibration. NeurIPS
  (2019)

\bibitem{lee2018simple}
Lee, K., Lee, K., Lee, H., Shin, J.: A simple unified framework for detecting
  out-of-distribution samples and adversarial attacks. In: NeurIPS. pp.
  7167--7177 (2018)

\bibitem{leidner2015classifying}
Leidner, D., Borst, C., Dietrich, A., Beetz, M., Albu-Sch{\"a}ffer, A.:
  Classifying compliant manipulation tasks for automated planning in robotics.
  In: 2015 IEEE/RSJ International Conference on Intelligent Robots and Systems
  (IROS). pp. 1769--1776 (2015)

\bibitem{LiangLS18}
Liang, S., Li, Y., Srikant, R.: Enhancing the reliability of
  out-of-distribution image detection in neural networks. In: ICLR (2018)

\bibitem{LinGGHD20}
Lin, T., Goyal, P., Girshick, R.B., He, K., Doll{\'{a}}r, P.: Focal loss for
  dense object detection. {IEEE} Trans. Pattern Anal. Mach. Intell. pp.
  318--327 (2020)

\bibitem{Liu2022CVPR}
Liu, B., Ben~Ayed, I., Galdran, A., Dolz, J.: The devil is in the margin:
  Margin-based label smoothing for network calibration. In: CVPR. pp. 80--88
  (June 2022)

\bibitem{luo2021learning}
Luo, Y., Wong, Y., Kankanhalli, M.S., Zhao, Q.: Learning to predict
  trustworthiness with steep slope loss. NeurIPS  (2021)

\bibitem{maddox2019simple}
Maddox, W.J., Izmailov, P., Garipov, T., Vetrov, D.P., Wilson, A.G.: A simple
  baseline for bayesian uncertainty in deep learning. NeurIPS  \textbf{32}
  (2019)

\bibitem{minderer2021revisiting}
Minderer, M., Djolonga, J., Romijnders, R., Hubis, F., Zhai, X., Houlsby, N.,
  Tran, D., Lucic, M.: Revisiting the calibration of modern neural networks.
  NeurIPS  (2021)

\bibitem{miotto2016deep}
Miotto, R., Li, L., Kidd, B.A., Dudley, J.T.: Deep patient: an unsupervised
  representation to predict the future of patients from the electronic health
  records. Scientific reports p. 26094 (2016)

\bibitem{MoonKSH20}
Moon, J., Kim, J., Shin, Y., Hwang, S.: Confidence-aware learning for deep
  neural networks. In: ICML. pp. 7034--7044 (2020)

\bibitem{Mozafari2019UnsupervisedTS}
Mozafari, A.S., Gomes, H.S., Le{\~a}o, W., Gagn{\'e}, C.: Unsupervised
  temperature scaling: An unsupervised post-processing calibration method of
  deep networks. arXiv: Computer Vision and Pattern Recognition  (2019)

\bibitem{MukhotiKSGTD20}
Mukhoti, J., Kulharia, V., Sanyal, A., Golodetz, S., Torr, P.H.S., Dokania,
  P.K.: Calibrating deep neural networks using focal loss. In: NeurIPS (2020)

\bibitem{muller2019does}
M{\"u}ller, R., Kornblith, S., Hinton, G.: When does label smoothing help? In:
  NeurIPS. pp. 4696--4705 (2019)

\bibitem{pml1Book}
Murphy, K.P.: Probabilistic Machine Learning: An introduction. MIT Press
  (2022), \url{probml.ai}

\bibitem{Naeini2015ObtainingWC}
Naeini, M.P., Cooper, G.F., Hauskrecht, M.: Obtaining well calibrated
  probabilities using bayesian binning. In: AAAI. pp. 2901--2907 (2015)

\bibitem{nixon2019measuring}
Nixon, J., Dusenberry, M.W., Zhang, L., Jerfel, G., Tran, D.: Measuring
  calibration in deep learning. In: CVPR Workshops. vol.~2 (2019)

\bibitem{ovadia2019can}
Ovadia, Y., Fertig, E., Ren, J., Nado, Z., Sculley, D., Nowozin, S., Dillon,
  J., Lakshminarayanan, B., Snoek, J.: Can you trust your model's uncertainty?
  evaluating predictive uncertainty under dataset shift. NeurIPS  (2019)

\bibitem{patel2020multi}
Patel, K., Beluch, W.H., Yang, B., Pfeiffer, M., Zhang, D.: Multi-class
  uncertainty calibration via mutual information maximization-based binning.
  In: ICLR (2020)

\bibitem{pereyra2017regularizing}
Pereyra, G., Tucker, G., Chorowski, J., Kaiser, {\L}., Hinton, G.: Regularizing
  neural networks by penalizing confident output distributions. arXiv preprint
  arXiv:1701.06548  (2017)

\bibitem{pittorino2021entropic}
Pittorino, F., Lucibello, C., Feinauer, C., Perugini, G., Baldassi, C.,
  Demyanenko, E., Zecchina, R.: Entropic gradient descent algorithms and wide
  flat minima. Journal of Statistical Mechanics: Theory and Experiment (12),
  124015 (2021)

\bibitem{RahimiSC0B20}
Rahimi, A., Shaban, A., Cheng, C., Hartley, R., Boots, B.: Intra
  order-preserving functions for calibration of multi-class neural networks.
  In: NeurIPS (2020)

\bibitem{rice2020overfitting}
Rice, L., Wong, E., Kolter, Z.: Overfitting in adversarially robust deep
  learning. In: ICML. pp. 8093--8104 (2020)

\bibitem{ShehzadBFARKK20}
Shehzad, M.N., Bashir, Q., Farooq, U., Ahmed, G., Raza, M., Kumar, P.M.,
  Khalid, M.: Threshold temperature scaling: Heuristic to address temperature
  and power issues in mpsocs. Microprocess. Microsystems p. 103124 (2020)

\bibitem{shen2020label}
Shen, Z., Liu, Z., Xu, D., Chen, Z., Cheng, K.T., Savvides, M.: Is label
  smoothing truly incompatible with knowledge distillation: An empirical study.
  In: ICLR (2020)

\bibitem{SimonyanZ14a}
Simonyan, K., Zisserman, A.: Very deep convolutional networks for large-scale
  image recognition. In: ICLR (2015)

\bibitem{tan2019efficientnet}
Tan, M., Le, Q.: Efficientnet: Rethinking model scaling for convolutional
  neural networks. In: ICML. pp. 6105--6114 (2019)

\bibitem{thulasidasan2019mixup}
Thulasidasan, S., Chennupati, G., Bilmes, J., Bhattacharya, T., Michalak, S.:
  On mixup training: Improved calibration and predictive uncertainty for deep
  neural networks. In: NeurIPS. pp. 13888--13899 (2019)

\bibitem{tolstikhin2021mlp}
Tolstikhin, I.O., Houlsby, N., Kolesnikov, A., Beyer, L., Zhai, X.,
  Unterthiner, T., Yung, J., Steiner, A., Keysers, D., Uszkoreit, J., et~al.:
  Mlp-mixer: An all-mlp architecture for vision. NeurIPS  (2021)

\bibitem{trockman2022patches}
Trockman, A., Kolter, J.Z.: Patches are all you need? arXiv preprint
  arXiv:2201.09792  (2022)

\bibitem{vaicenavicius2019evaluating}
Vaicenavicius, J., Widmann, D., Andersson, C., Lindsten, F., Roll, J.,
  Sch{\"o}n, T.: Evaluating model calibration in classification. In: AISTATS.
  pp. 3459--3467 (2019)

\bibitem{RethinkingCalibration}
Wang, D., Feng, L., Zhang, M.: Rethinking calibration of deep neural networks:
  Do not be afraid of overconfidence. In: NeurIPS (2021)

\bibitem{wen2020combining}
Wen, Y., Jerfel, G., Muller, R., Dusenberry, M.W., Snoek, J., Lakshminarayanan,
  B., Tran, D.: Combining ensembles and data augmentation can harm your
  calibration. In: ICLR (2020)

\bibitem{WuX020}
Wu, D., Xia, S., Wang, Y.: Adversarial weight perturbation helps robust
  generalization. In: NeurIPS (2020)

\bibitem{XingAZP20}
Xing, C., Arik, S.{\"{O}}., Zhang, Z., Pfister, T.: Distance-based learning
  from errors for confidence calibration. In: ICLR (2020)

\bibitem{yao2015tiny}
Yao, L., Miller, J.: Tiny imagenet classification with convolutional neural
  networks. CS 231N

\bibitem{YunPLS20}
Yun, S., Park, J., Lee, K., Shin, J.: Regularizing class-wise predictions via
  self-knowledge distillation. In: CVPR. pp. 13873--13882 (2020)

\bibitem{zagoruyko2016wide}
Zagoruyko, S., Komodakis, N.: Wide residual networks. In: BMVC (2016)

\bibitem{zhang2018mixup}
Zhang, H., Cisse, M., Dauphin, Y.N., Lopez-Paz, D.: Mixup: Beyond empirical
  risk minimization. In: ICLR (2018)

\bibitem{howmixup}
Zhang, L., Deng, Z., Kawaguchi, K., Zou, J.: When and how mixup improves
  calibration. In: ICML. pp. 26135--26160 (2022)

\bibitem{mixCalibration}
Zhang, W., Vaidya, I.: Mixup training leads to reduced overfitting and improved
  calibration for the transformer architecture. CoRR  (2021)

\bibitem{zhong2021improving}
Zhong, Z., Cui, J., Liu, S., Jia, J.: Improving calibration for long-tailed
  recognition. In: CVPR. pp. 16489--16498 (2021)

\bibitem{zhu2022learning}
Zhu, F., Zhang, X.Y., Wang, R.Q., Liu, C.L.: Learning by seeing more classes.
  IEEE Transactions on Pattern Analysis and Machine Intelligence  (2022)

\end{thebibliography}

\appendix
\onecolumn
\clearpage
\begin{center}{\bf {\Large \emph{Supplementary Materials}:}}
\end{center}
\begin{center}{\bf {\Large Rethinking Confidence Calibration for Failure Prediction}}
\end{center}
\vspace{0.1in}

\section{Evaluation Metrics}
\subsection{Failure Prediction}
\setParDis
\textbf{AURC \& E-AURC.} AURC measures the area under the
curve drawn by plotting the risk according to coverage. The
coverage indicates the ratio of samples whose confidence
estimates are higher than some confidence threshold, and the
risk, also known as the selective risk \cite{GeifmanE17}, is an error rate computed by using those samples. A low value of AURC implies that correct and incorrect predictions can be well-separable by confidence estimates
associated with samples.
Inherently, AURC is affected by the predictive performance
of a model. To have a unitless performance measure that can
be applied across models, Geifman \emph{et al.}, \cite{GeifmanUE19} introduce
a normalized AURC, named Excess-AURC (E-AURC). Specifically, E-AURC
can be computed by subtracting the optimal AURC,
the lowest possible value for a given model, from the empirical
AURC.

\noindent
\textbf{FPR-95\%TPR.} FPR-95\%TPR can be interpreted as the probability that a negative (misclassified) example is predicted as a correct one when the true positive rate (TPR) is as high as $95\%$. True positive rate can be computed by TPR=TP/(TP+FN), where TP and FN denote the number of true positives and false negatives, respectively. The false positive rate (FPR) can be computed by FPR=FP/(FP+TN), where FP and TN denote the number of false positives and true negatives, respectively.

\noindent
\textbf{AUROC.} 
AUROC measures the area under the receiver operating characteristic curve. The ROC curve depicts the relationship between true positive rate and false positive rate. This metric is a threshold-independent performance evaluation. The AUROC can be interpreted as the probability that a positive example is assigned a higher prediction score than a negative example.

\noindent
\textbf{AUPR-Success \& AUPR-Error.}
AUPR is the area under the precision-recall curve. The precision-recall curve is a graph showing the precision=TP/(TP+FP) versus recall=TP/(TP+FN). The metrics AUPR-Success and AUPR-Error indicate the area under the precision-recall curve where correct predictions and errors are specified as positives, respectively.

\subsection{Confidence Calibration}
\noindent
\textbf{ECE.} 
Confidence calibration aims to narrow the mismatch between a model's confidence and its accuracy. 
As an approximation of such difference, Expected calibration error (ECE) \cite{Naeini2015ObtainingWC} bins the predictions in $[0,1]$ under $M$ euqally-spaced intervals, and then averages the accuracy/confidence in each bin. Then the ECE can be computed as
\begin{equation}
\begin{aligned}
\text{ECE} = \sum_{m=1}^M \frac{\left|B_{m}\right|}{n} \left|\text{acc}(B_{m})-\text{avgConf}(B_{m})\right|,
\end{aligned}
\end{equation}
where $n$ is the number of all samples. 

\noindent
\textbf{NLL.}
Negative log likelihood (NLL) is a standard measure of a probabilistic model's quality, which is defined as
\begin{equation}
\text{NLL} = -\sum_{i=1}^n \text{log}[\hat{p}(y_{c}|\bm{x_{i}})],
\end{equation}
where $y_{c}$ donates the element for ground-truth class.
In expectation, NLL is minimized if and only if $\hat{p}(Y|\bm{X})$ recovers the truth conditional distribution.

\noindent
\textbf{Brier Score.}
Brier score \cite{Brier1950} can be interpreted
as the average mean squared error between the predicted
probability and one-hot encoded label. It can be computed
as
\begin{equation}
\text{Brier} = \frac{1}{n}\sum_{i=1}^n\sum_{k=1}^K [\hat{p}(y_{k}|\bm{x_{i}})-t_{k}],
\end{equation}
where $t_{k}=1$ if $k = c$ (ground-truth class), and 0 otherwise.
\setParDef

\section{Details of Calibration Methods}
\setParDis
\textbf{Mixup.} Mixup \cite{zhang2018mixup} trains a model on convex combinations of pairs of examples and their labels to encourage linear interpolating predictions. Given a pair of examples $(\bm{x}_{a}, \bm{y}_{a})$ and $(\bm{x}_{b}, \bm{y}_{b})$ sampled from the mini-batch, where $\bm{x}_{a}, \bm{x}_{b}$ represent different samples and $\bm{y}_{a}, \bm{y}_{b}$ denote their one-hot label vectors. Mixup applies linear interpolation to produce augmented data $(\widetilde{\bm{x}}, \widetilde{\bm{y}})$ as follows:
\begin{equation}
\begin{aligned}
\widetilde{\bm{x}} = \lambda \bm{x}_{a} + (1-\lambda) \bm{x}_{b}, ~
\widetilde{\bm{y}} = \lambda \bm{y}_{a} + (1-\lambda) \bm{y}_{b}.
\end{aligned}
\end{equation}
The $\lambda \in [0,1]$ is a random parameter sampled as $\lambda\sim\text{Beta}(\alpha, \alpha)$ for $\alpha \in (0, \infty)$. Thulasidasan \emph{et~al.} \cite{thulasidasan2019mixup} empirically found that mixup can significantly improve confidence calibration of DNNs. Similar calibration effect of mixup has been verified in natural language processing tasks \cite{mixCalibration}. 

\noindent\textbf{Label Smoothing.} Label Smoothing (LS) is commonly used as regularization to reduce overfitting of DNNs. Specifically, when training the model with empirical risk minimization, the one-hot label $\bm{y}$ (\emph{i.e.} the element $y_c$ is 1 for ground-truth class and 0 for others) is smoothed by distributing 
a fraction of mass over the other non ground-truth classes:
\begin{equation}
\widetilde{\bm{y}_i} = \left\{ 
\begin{aligned}
1-\epsilon, &~\text{if} ~~i=c, \\
\epsilon/(K-1), &~\text{otherwise}.
\end{aligned}
\right.
\end{equation}
where $\epsilon$ is a small positive constant coefficient for smoothing the one-hot label, and $K$ is the number of training classes. Recently, Muller \emph{et al}. \cite{muller2019does, thulasidasan2019mixup} showed the favorable calibration effect of LS. 

\begin{table}[!t]
	\vskip 0.05in
	\caption{Evaluating calibration methods for failure prediction on MobileNet and EfficientNet. AURC and E-AURC values are multiplied by $10^3$ for clarity, and all remaining values are percentage.}
	\vskip -0.2in
	\label{table-s1}
	\begin{center}
		\renewcommand\tabcolsep{7.3pt}
		\begin{small}
			\newcommand{\tabincell}[2]{\begin{tabular}{@{}#1@{}}#2\end{tabular}}
			\scalebox{0.69}{
				\renewcommand{\arraystretch}{1.1}
				\begin{tabular}{llcccccc}
					\toprule
					\multicolumn{8}{c}{\textbf{CIFAR-10}}  \\ \midrule
					\textbf{Network} 
					& \textbf{Method}&
					\tabincell{c}{\textbf{AURC} \\ \textbf{($\downarrow$)}} & \tabincell{c}{\textbf{E-AURC} \\ \textbf{($\downarrow$)}}&
					\tabincell{c}{\textbf{FPR-95\%} \\ \textbf{TPR($\downarrow$)}} &
					\tabincell{c}{\textbf{AUROC} \\ \textbf{($\uparrow$)}} & \tabincell{c}{\textbf{AUPR-} \\ \textbf{Success($\uparrow$)}} & \tabincell{c}{\textbf{AUPR-} \\ \textbf{Error($\uparrow$)}}\\
					\midrule
					\multirow{6}{*}{MobileNet-v2}
					& baseline \cite{hendrycks2017baseline}  &\bftab{9.65$\pm$0.09} & \bftab{6.71$\pm$0.15} & \bftab{45.62$\pm$0.69} & \bftab{92.29$\pm$0.18} & \bftab{99.29$\pm$0.02} &\bftab{46.98$\pm$0.61} \\
					& mixup \cite{thulasidasan2019mixup}  &10.38$\pm$0.13 & 8.06$\pm$0.17 & 47.25$\pm$0.35 & 90.86$\pm$0.22 & 99.14$\pm$0.02 & 42.44$\pm$0.87 \\
					& LS \cite{muller2019does}  &14.49$\pm$0.59 & 11.84$\pm$0.63 & 46.21$\pm$2.47 & 88.93$\pm$0.30 & 98.74$\pm$0.07 & 43.55$\pm$1.69 \\
					& Focal \cite{MukhotiKSGTD20}  &11.59$\pm$0.40 & 8.04$\pm$0.22 & 48.91$\pm$0.30 & 91.44$\pm$0.17 & 99.13$\pm$0.03 & 45.30$\pm$0.60 \\
					& CS-KD \cite{YunPLS20}  &23.50$\pm$1.63 & 18.32$\pm$1.46 & 52.82$\pm$0.60 & 87.17$\pm$0.66 & 98.00$\pm$0.16 & 46.20$\pm$1.39 \\
					& L1 \cite{joo2020revisiting}  &11.26$\pm$0.48 & 8.74$\pm$0.43 & 47.10$\pm$2.21 & 90.36$\pm$0.39 & 99.07$\pm$0.05 & 43.64$\pm$1.97 \\
					\midrule
					
					\multirow{6}{*}{EfficientNet}
					& baseline \cite{hendrycks2017baseline} & \bftab{16.24$\pm$0.32} & \bftab{11.24$\pm$0.45} & 53.41$\pm$1.00 & \bftab{90.23$\pm$0.19} & \bftab{98.78$\pm$0.05} & \bftab{47.88$\pm$1.75} \\
					& mixup \cite{thulasidasan2019mixup}  & 16.24$\pm$2.27 & 11.85$\pm$2.12 & 52.25$\pm$1.01 & 90.11$\pm$0.75 & 98.72$\pm$0.23 & 46.97$\pm$0.84 \\
					& LS \cite{muller2019does}  & 24.56$\pm$0.72 & 19.91$\pm$0.70 & 51.39$\pm$0.70 & 87.72$\pm$0.21 & 97.84$\pm$0.07 & 47.08$\pm$0.34 \\
					& Focal \cite{MukhotiKSGTD20}  & 18.93$\pm$0.40 & 13.39$\pm$0.36 & 55.06$\pm$1.48 & 89.24$\pm$0.50 & 98.54$\pm$0.04 & 47.79$\pm$1.74 \\
					& CS-KD \cite{YunPLS20} & 21.15$\pm$0.90 & 16.77$\pm$1.03 & 50.93$\pm$2.09 & 88.05$\pm$0.74 & 98.18$\pm$0.11 & 46.57$\pm$1.79 \\
					& L1 \cite{joo2020revisiting}  & 20.21$\pm$1.04 & 15.62$\pm$1.28 & \bftab{51.16$\pm$0.61} & 88.97$\pm$0.55 & 98.30$\pm$0.14 & 48.26$\pm$1.24 \\
					
					\bottomrule
					\toprule
					\multicolumn{8}{c}{\textbf{CIFAR-100}}  \\ \midrule
					\textbf{Network} 
					& \textbf{Method}&
					\tabincell{c}{\textbf{AURC} \\ \textbf{($\downarrow$)}} & \tabincell{c}{\textbf{E-AURC} \\ \textbf{($\downarrow$)}}&
					\tabincell{c}{\textbf{FPR-95\%} \\ \textbf{TPR($\downarrow$)}} &
					\tabincell{c}{\textbf{AUROC} \\ \textbf{($\uparrow$)}} & \tabincell{c}{\textbf{AUPR-} \\ \textbf{Success($\uparrow$)}} & \tabincell{c}{\textbf{AUPR-} \\ \textbf{Error($\uparrow$)}}\\
					\midrule
					\multirow{6}{*}{MobileNet-v2}
					& baseline \cite{hendrycks2017baseline}  &\bftab{90.02$\pm$3.39} & \bftab{46.06$\pm$2.85} & \bftab{65.61$\pm$1.57} & \bftab{85.96$\pm$0.71} & 94.05$\pm$0.37 & \bftab{66.96$\pm$1.39} \\
					& mixup \cite{thulasidasan2019mixup}  & 93.78$\pm$0.81 & 46.77$\pm$1.14 & 67.60$\pm$1.08 & 85.60$\pm$0.53 & \bftab{94.70$\pm$0.39} & 64.44$\pm$1.00 \\
					& LS \cite{muller2019does}  & 92.61$\pm$0.90 & 49.07$\pm$0.51 & 67.18$\pm$0.90 & 84.98$\pm$0.12 & 93.67$\pm$0.07 & 65.86$\pm$0.86 \\
					& Focal \cite{MukhotiKSGTD20}  & 103.66$\pm$2.21 & 53.26$\pm$1.38 & 68.95$\pm$1.05 & 84.31$\pm$0.26 & 93.01$\pm$0.05 & 65.47$\pm$0.61\\
					& CS-KD \cite{YunPLS20}  & 121.31$\pm$4.54 & 60.59$\pm$3.17 & 66.45$\pm$1.25 & 84.77$\pm$0.31 & 91.73$\pm$0.48 & 65.16$\pm$0.24 \\
					& L1 \cite{joo2020revisiting}  & 90.07$\pm$0.91 & 46.93$\pm$0.77 & 66.79$\pm$1.76 & 85.44$\pm$0.32 & 93.96$\pm$0.09 & 66.10$\pm$0.78 \\
					
					\midrule
					\multirow{6}{*}{EfficientNet}
					& baseline \cite{hendrycks2017baseline}  &  \bftab{109.88$\pm$1.58} & \bftab{55.06$\pm$0.21} & 67.29$\pm$0.91 & \bftab{85.18$\pm$0.24} & \bftab{92.77$\pm$0.02} & \bftab{70.95$\pm$0.78} \\
					& mixup \cite{thulasidasan2019mixup}  & 114.19$\pm$3.13 & 57.14$\pm$1.69 & 67.06$\pm$2.20 & 84.71$\pm$0.34 & 92.31$\pm$0.25 & 68.92$\pm$0.76 \\
					& LS \cite{muller2019does}  & 116.48$\pm$2.63 & 59.73$\pm$1.38 & \bftab{65.84$\pm$1.28} & 84.64$\pm$0.09 & 91.94$\pm$0.22 & 69.48$\pm$0.23 \\
					& Focal \cite{MukhotiKSGTD20}  & 132.34$\pm$2.60 & 65.47$\pm$3.31 & 69.01$\pm$1.77 & 83.51$\pm$0.71 & 90.97$\pm$0.44 & 69.20$\pm$0.88 \\
					& CS-KD \cite{YunPLS20}  &117.75$\pm$0.86 & 55.60$\pm$0.80 & 66.97$\pm$1.16 & 84.90$\pm$0.44 & 92.61$\pm$0.09 & 68.23$\pm$0.65 \\
					& L1 \cite{joo2020revisiting}  & 113.47$\pm$2.77 & 58.56$\pm$1.28 & 67.32$\pm$1.32 & 84.33$\pm$0.32 & 92.16$\pm$0.20 & 67.87$\pm$0.54 \\
					
					\bottomrule
			\end{tabular}}
		\end{small}
	\end{center}
	\vskip -0.3in
\end{table}

\noindent\textbf{Focal Loss.} Focal Loss \cite{LinGGHD20} modifies the standard cross entropy loss by weighting loss components of samples in a mini-batch according to how well the model classifies them: $\mathcal{L}_f := -(1-\hat{p}_{i,y_i})^\gamma ~\text{log}~\hat{p}_{i,y_i}$, where $\gamma$ is a strength coefficient. Intuitively, with focal loss, the gradients of correctly classified samples are restrained and those of incorrectly classified samples are emphasized. Mukhoti \emph{et~al}. \cite{MukhotiKSGTD20} demonstrated that focal loss can automatically learn well-calibrated models.

\noindent\textbf{CS-KD.} Class-wise self-knowledge distillation (CS-KD) \cite{YunPLS20} method alleviates the overfitting problem of DNNs by penalizing the predictive distribution between the samples within the same class:
\begin{equation}
\begin{aligned}
\mathcal{L}_{CS-KD}(\bm{x}, \bm{x'},y,T) := \mathcal{L}_{CE}(\bm{x},y) + \lambda_{cls} \cdot T^2 \cdot \text{KL}(P(y|\bm{x'};T)\Vert P(y|\bm{x};T)),
\end{aligned}
\end{equation}
where KL denotes the Kullback-Leibler (KL) divergence, $\mathcal{L}_{CE}$ is the standard cross-entropy loss, $T$ is the temperature and $\lambda_{cls}$ is a loss weight for the class-wise regularization. Particularly, Yun \emph{et al.} \cite{YunPLS20} reported the positive effectiveness of CS-KD on confidence calibration. 

\noindent\textbf{$L_p$ Norm.} Recently, Joo \emph{et~al}. \cite{joo2020revisiting} explored the effect of explicit regularization strategies (\emph{e.g.}, $L_p$ norm in the logits space) for calibration. Specifically, the learning objective is:
\begin{equation}
\begin{aligned}
\mathcal{L}_{L_P}(\bm{x}, y) := \mathcal{L}_{CE}(\bm{x},y) + \lambda \Vert f(\bm{x})\Vert,
\end{aligned}
\end{equation}
where $f(\bm{x})$ donates the logit of $\bm{x}$, and $\lambda$ is a strength coefficient. Although being simple, $L_p$ norm (\emph{e.g.}, $L_1$ norm) can provide well-calibrated predictive uncertainty \cite{joo2020revisiting}. 
\setParDef

\begin{figure*}[t]
	\begin{center}
		\vskip -0.05 in
		\centerline{\includegraphics[width=0.88\textwidth]{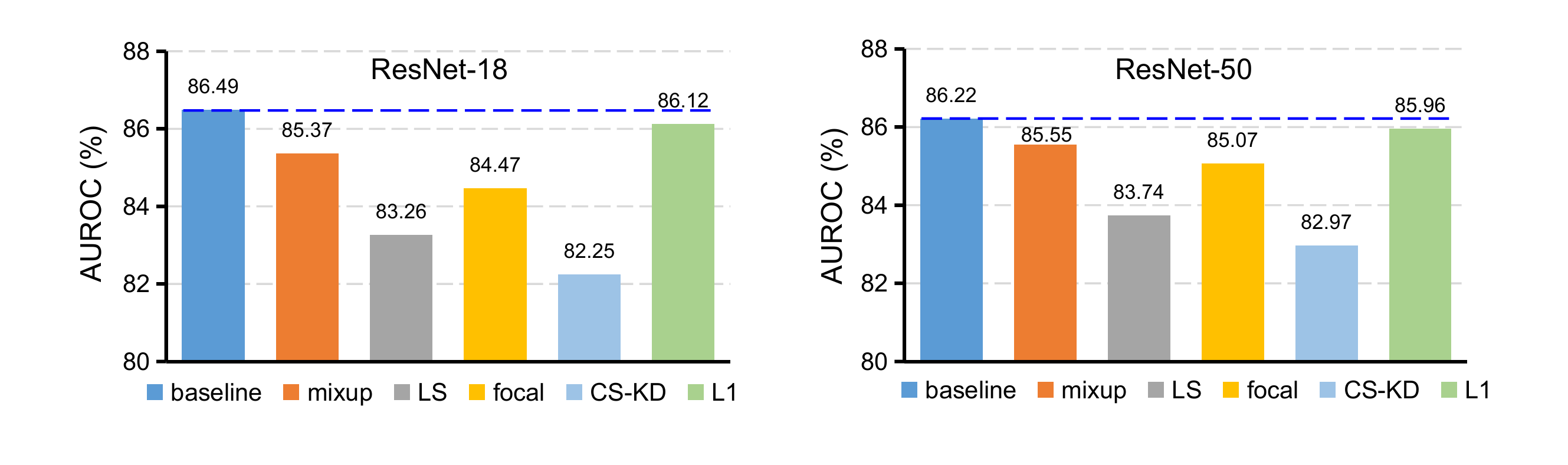}}
		\vskip -0.23 in
		\caption{Large-scale experiments on Tiny-ImageNet with ResNet-18 and ResNet-50.}
		\label{figure-s1}
	\end{center}
	\vskip -0.3 in
\end{figure*}

\begin{figure*}[t]
	\begin{center}
		\vskip -0.1 in
		\centerline{\includegraphics[width=0.9\textwidth]{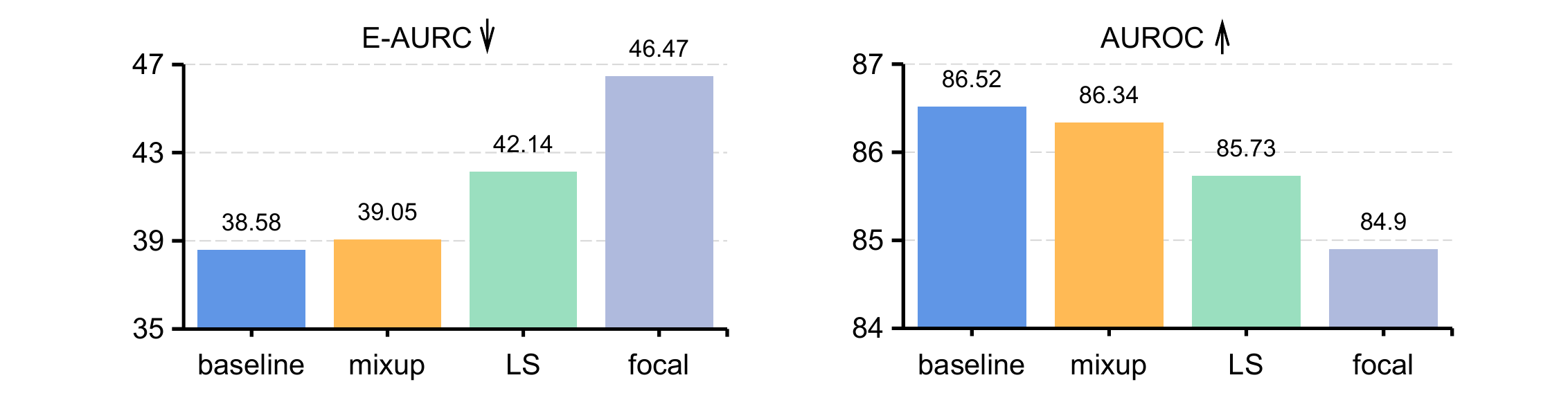}}
		\vskip -0.17 in
		\caption{Large-scale experiments on ImageNet with ResNet-50.}
		\label{figure-s2}
	\end{center}
	\vskip -0.3 in
\end{figure*}
\section{Experiments}
\subsection{Evaluating Calibration Methods for Failure Prediction}

\textbf{Hyper-parameter setup of calibration methods.} In our experiments, for mixup, we follow the setting in \cite{thulasidasan2019mixup} to use $\alpha=0.2$. For LS, it has been shown that an $\epsilon \in [0.05, 0.1]$ performs best for calibration. Therefore, $\epsilon = 0.05$ is used in our experiments. For focal loss, we set the hyperparameter $\gamma=3$ following the suggestion in \cite{MukhotiKSGTD20}. For CS-KD, we use the default setting of hyper-parameters for implementation based on their open-sourced code \url{https://github.com/alinlab/cs-kd}. For $L_p$ norm, we use the $L_1$ norm, which is effective for calibration, as shown in \cite{joo2020revisiting}, and $\lambda=0.01$ is used in our experiments. The details of the calibration methods and definition of those hyper-parameters can be found in Section B.

\setParDis
\noindent\textbf{Results on MobileNet and EfficientNet.} We provide the results of more networks: MobileNet-v2 \cite{howard2017mobilenets} and EfficientNet \cite{tan2019efficientnet} in Table~\ref{table-s1}, from which we can observe similar negative effect of calibration methods on failure prediction. 

\noindent\textbf{Results on Tiny-ImageNet.} For experiments on Tiny-ImageNet \cite{yao2015tiny}, the models (ResNet-18 and ResNet-50) are trained from sketch using SGD with a momentum of 0.9, an initial learning rate of 0.1, and a weight decay of 5e-4 for 90 epochs with the mini-batch size of 128. The learning rate is reduced by a factor of 10 at 40, and 70 epochs. The results are shown in Fig.~\ref{figure-s1}. As can be seen, baseline has the highest AUROC values, which indicates that baseline still performs better than those compared calibration methods for failure prediction.

\noindent\textbf{Results on ImageNet with ResNet-50.} We perform the automatic mixed precision training using the open-sourced code \url{https://github.com/NVIDIA/apex/tree/master/examples/imagenet}. The results of on ImageNet dataset with ResNet-50 are shown in Fig.~\ref{figure-s2}, from which we can observe similar negative effect of calibration methods on failure prediction.
\setParDef

\begin{algorithm}[!t]
	\caption{FMFP: Flat Minima for Failure Prediction algorithm}
	\KwIn{Model Weights $\bm{\theta}$, scheduled learning $\alpha$, cycle length $c$, number of iterations $K$, averaging start epoch $S$, neighborhood size $\rho$, the number of past checkpoints to be averaged $n$, loss function $\mathcal{L}$}
	\KwOut{Model trained with FMFP}
	\For{i $\leftarrow$ 1 to K}{ 
		Sample a mini-batch data\\
		Compute gradient $\nabla \mathcal{L(\bm{\theta})}$ of the batch’s training loss\\
		Compute worst-case perturbation $\hat{\epsilon} \leftarrow \rho \frac{\nabla \mathcal{L(\bm{\theta})}}{\Vert \nabla \mathcal{L(\bm{\theta})}\Vert_2} $\\
		Gradient update $\bm{\theta} \leftarrow \bm{\theta} - \alpha \nabla \mathcal{L(\bm{\theta} + \hat{\epsilon})}$\\
		
		\eIf{$i \ge S$ and mod$(i, c)=0$} {
			$\bm{\theta}_{sswa}^t \leftarrow \frac{{\bm{\theta}_{sswa}}^{t-1} \times n + \bm{\theta}^t}{n+1}$
		}{
			i ++ \\
		}
	}	
\end{algorithm}

\begin{table}[t]
	\caption{Confidence estimation on Tiny-ImageNet dataset. The means and standard deviations
		over three runs are reported. AURC and E-AURC values are multiplied by $10^3$, and NLL are multiplied by 10 for clarity. Remaining values are percentages.}
	\vskip -0.25in
	\label{table-s2}
	\begin{center}
		\renewcommand\tabcolsep{3pt}
		\begin{small}
			\newcommand{\tabincell}[2]{\begin{tabular}{@{}#1@{}}#2\end{tabular}}
			\scalebox{0.67}{
				\renewcommand{\arraystretch}{1.2}
				\begin{tabular}{llcccccccc}
					\toprule
					\textbf{Network} 
					& \textbf{Method}&
					\tabincell{c}{\textbf{AURC} \\ \textbf{($\downarrow$)}} & \tabincell{c}{\textbf{E-AURC} \\ \textbf{($\downarrow$)}}&
					\tabincell{c}{\textbf{FPR-95\%} \\ \textbf{TPR($\downarrow$)}} &
					\tabincell{c}{\textbf{AUROC} \\ \textbf{($\uparrow$)}} & \tabincell{c}{\textbf{AUPR-} \\ \textbf{Success($\uparrow$)}} & \tabincell{c}{\textbf{AUPR-} \\ \textbf{Error($\uparrow$)}} &
					\tabincell{c}{\textbf{ECE} \\ \textbf{($\downarrow$)}} & \tabincell{c}{\textbf{NLL} \\ \textbf{($\downarrow$)}}\\
					\midrule
					\multirow{3}{*}{ResNet-18}
					& baseline \cite{hendrycks2017baseline}  & 124.54$\pm$0.97 & 53.13$\pm$0.12 & 62.45$\pm$0.68 & 86.49$\pm$0.11 & 92.54$\pm$0.04 & 75.10$\pm$0.57 & 10.13$\pm$0.42 & 15.76$\pm$0.14 \\
					& CRL \cite{MoonKSH20}  & 118.05$\pm$1.88 & 49.55$\pm$0.75 & \bftab{60.65$\pm$1.85} & 86.62$\pm$0.37 & 93.10$\pm$0.09 & \bftab{75.47$\pm$1.30} & 7.56$\pm$0.64 & 14.69$\pm$0.09 \\ 
					& ours  & \bftab{107.01$\pm$1.17} & \bftab{46.88$\pm$0.89} & 62.35$\pm$1.38 & \bftab{86.86$\pm$0.27} & \bftab{93.65$\pm$0.11} & 73.14$\pm$0.75 & \bftab{4.79$\pm$0.20} & \bftab{13.17$\pm$0.05} \\
					
					\midrule
					\multirow{3}{*}{ResNet-50}
					& baseline \cite{hendrycks2017baseline}  & 119.01$\pm$3.19 & 53.05$\pm$1.04 & 63.56$\pm$0.20 & 86.22$\pm$0.20 & 92.66$\pm$0.18 & 73.74$\pm$0.61 & 10.14$\pm$0.10 & 15.41$\pm$0.26 \\
					& CRL \cite{MoonKSH20}  & 110.80$\pm$2.33 & 48.27$\pm$1.21 & 62.01$\pm$0.35 & 87.02$\pm$0.18 & 93.38$\pm$0.19 & 74.15$\pm$0.26 & 7.35$\pm$0.29 & 14.23$\pm$0.19 \\ 
					& ours  & \bftab{98.43$\pm$1.17} & \bftab{42.55$\pm$0.89} & \bftab{60.71$\pm$1.38} & \bftab{87.71$\pm$0.27} & \bftab{94.28$\pm$0.11} & \bftab{74.19$\pm$0.75} & \bftab{4.85$\pm$0.20} & \bftab{12.57$\pm$0.05} \\
					\bottomrule
			\end{tabular}}
		\end{small}
	\end{center}
	\vskip -0.3in
\end{table}

\subsection{Improving Failure Prediction by Finding Flat Minima}
\noindent\textbf{Pseudo-code for FMFP.} Algorithm 1 gives pseudo-code for the full FMFP algorithm, which can be implemented by a few lines of codes in pytorch.
\begin{figure*}[t]
	\begin{center}
		\centerline{\includegraphics[width=\textwidth]{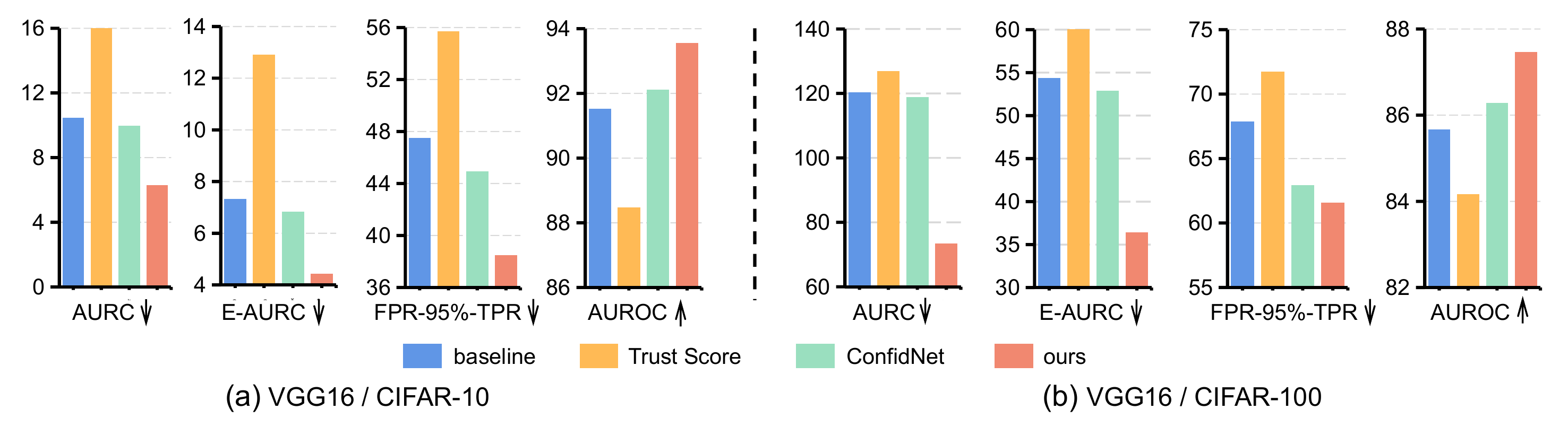}}
		\vskip -0.22 in
		\caption{Comparison with Trust Score and ConfidNet.}
		\label{figure-s4}
	\end{center}
	\vskip -0.3 in
\end{figure*}
\begin{figure*}[t]
	\begin{center}
		\centerline{\includegraphics[width=1.06\textwidth]{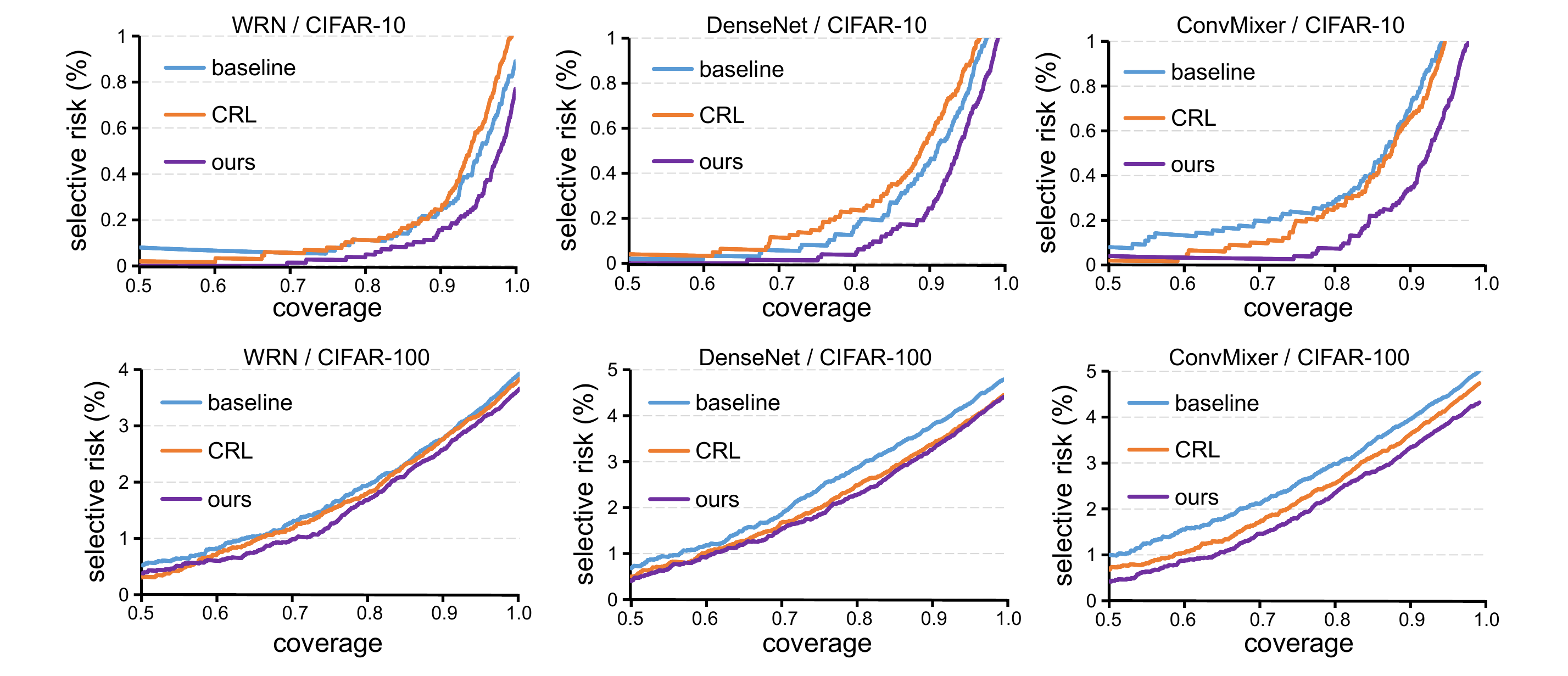}}
		\vskip -0.1 in
		\caption{Comparison of risk-coverage curves.}
		\label{figure-s5}
	\end{center}
	\vskip -0.4 in
\end{figure*}

\setParDis
\noindent\textbf{Implementation details.} For CRL \cite{MoonKSH20}, our implementation is based on open-sourced code \url{https://github.com/daintlab/confidence-aware-learning}. For SWA and FMFP, the cyclical learning rate schedule is used as suggested in \cite{izmailov2018averaging}.
For experiments on CIFAR-10 and CIFAR-100, checkpoints at 120-th epoch of the baseline
models are used as the initial point of SWA and FMFP. For experiments on Tiny-ImageNet, checkpoints at 50-th epoch of the baseline models are used as the initial point of SWA and FMFP. 

\noindent\textbf{Results on Tiny-ImageNet.} For experiments on Tiny-ImageNet \cite{yao2015tiny}, the models (ResNet-18 and ResNet-50) are trained from sketch using SGD with a momentum of 0.9, an initial learning rate of 0.1, and a weight decay of 5e-4 for 90 epochs with the mini-batch size of 128. The learning rate is reduced by a factor of 10 at 40, and 70 epochs. The results are shown in Table~\ref{table-s2}. We observe that ours method (FMFP) generally outperform the strong baseline and CRL on various metrics of failure prediction.

\noindent
\textbf{Comparison with ConfidNet.} ConfidNet \cite{corbiere2021confidence, corbiere2019addressing} is a failure prediction method that relies on misclassified samples in training set. Therefore, it can not be used for models with high training accuracy. Therefore, we make comparison on VGG network \cite{SimonyanZ14a} following the setting in \cite{corbiere2019addressing}. As reported in Fig.~\ref{figure-s4}, our method consistently outperforms ConfidNet under various metrics.

\noindent
\textbf{Selective risk-coverage curves results.} Fig.~\ref{figure-s5} plots more risk-coverage curves. As can be seen, our method yields lower risk at a given coverage.  

\noindent
\textbf{Relation with SWAG.} SWA-Gaussion (SWAG) \cite{maddox2019simple} is a bayesian inference technique, which is effective for calibration. We mainly differ from SWAG in the following three aspects. 
(1) \emph{Technique}: SWAG needs to sample many \emph{e.g.}, 100 times to obtain bayesian inference uncertainty while ours do not need. (2) \emph{Insight}: SWAG leverage the weight average property of SWA for bayesian approximate. We are motivated by the connection between flat minima and confidence separability, thus other techniques like SAM can also be used. (3) \emph{Problem setting}: SWAG focuses on calibration while we focus on failure prediction.
\setParDef

\end{document}